\documentclass[journal]{IEEEtran}
\usepackage{bbding}

\usepackage{color}

\newcommand{\revisionxin}[1]{\textcolor[rgb]{0,0,0}{#1}}

\usepackage{multicol}
\usepackage{multirow}

\ifCLASSINFOpdf
   \usepackage[pdftex]{graphicx}
\else
\fi
\usepackage{amssymb}

\usepackage{amsmath}
\begin{document}
%
\title{Night-time Scene Parsing with a Large Real Dataset}
%
%
%

\author{Xin Tan, \ \
        Ke Xu, \ \
        Ying Cao, \ \
        Yiheng Zhang, \ \
        Lizhuang Ma, \ \
        and \ \ Rynson W.H. Lau 
\thanks{X. Tan and K. Xu are with the Department of Computer Science and Engineering, Shanghai Jiao Tong University, Shanghai, China, and the Department of Computer Science, City University of Hong Kong, HKSAR, China.}
\thanks{L. Ma are with the Department of Computer Science and Engineering, Shanghai Jiao Tong University, Shanghai, China, and the School of Computer Science and Technology, East China Normal University, Shanghai, China.}
\thanks{Ying Cao and Rynson W.H. Lau are with the Department of Computer Science, City University of Hong Kong, HKSAR, China. }
\thanks{Y. Zhang is with the Department of Computer Science, Stanford University, CA, USA, and the Department of Computer Science and Engineering, Shanghai Jiao Tong University, Shanghai, China. }

\thanks{Manuscript received November 24, 2020; revised xx xx, 2021.}}

\markboth{Journal of \LaTeX\ Class Files,~Vol.~14, No.~8, August~2015}%
{Shell \MakeLowercase{\textit{et al.}}: Bare Demo of IEEEtran.cls for IEEE Journals}
%



\maketitle

\begin{abstract}
Although huge progress has been made on scene analysis in recent years, most existing works assume the input images to be in day-time with good lighting conditions. In this work, we aim to address the night-time scene parsing (NTSP) problem, which has two main challenges:
1) labeled night-time data are scarce, and 2) over- and under-exposures may co-occur in the input night-time images and are not explicitly modeled in existing pipelines. To tackle the scarcity of night-time data, we collect a novel labeled dataset, named {\it NightCity}, of 4,297 real night-time images with ground truth pixel-level semantic annotations. To our knowledge, NightCity is the largest dataset for NTSP. In addition, we also propose an exposure-aware framework to address the NTSP problem through augmenting the segmentation process with explicitly learned exposure features.
Extensive experiments show that training on NightCity can significantly improve NTSP performances and that our exposure-aware model outperforms the state-of-the-art methods, yielding top performances on our dataset as well as existing datasets.
\end{abstract}


%
\IEEEpeerreviewmaketitle

\section{Introduction}
\IEEEPARstart{S}{cene} Parsing is an important computer vision task for many applications, such as human parsing~\cite{Gong_2017_CVPR}, image editing~\cite{NEURIPS2018_653ac11c} and autonomous driving \cite{yang2018real}. Although a lot of methods have been proposed, they mainly focus on day-time scenes.
However, as night time and day time cover roughly about  50\% of  the time each (averaged over a year), it is equally important to build vision systems that perform well at night time, particularly for autonomous driving at night.
In this paper, we address the night-time scene parsing (NTSP) problem.

\begin{figure}[t]
\centering
\includegraphics[width=\linewidth]{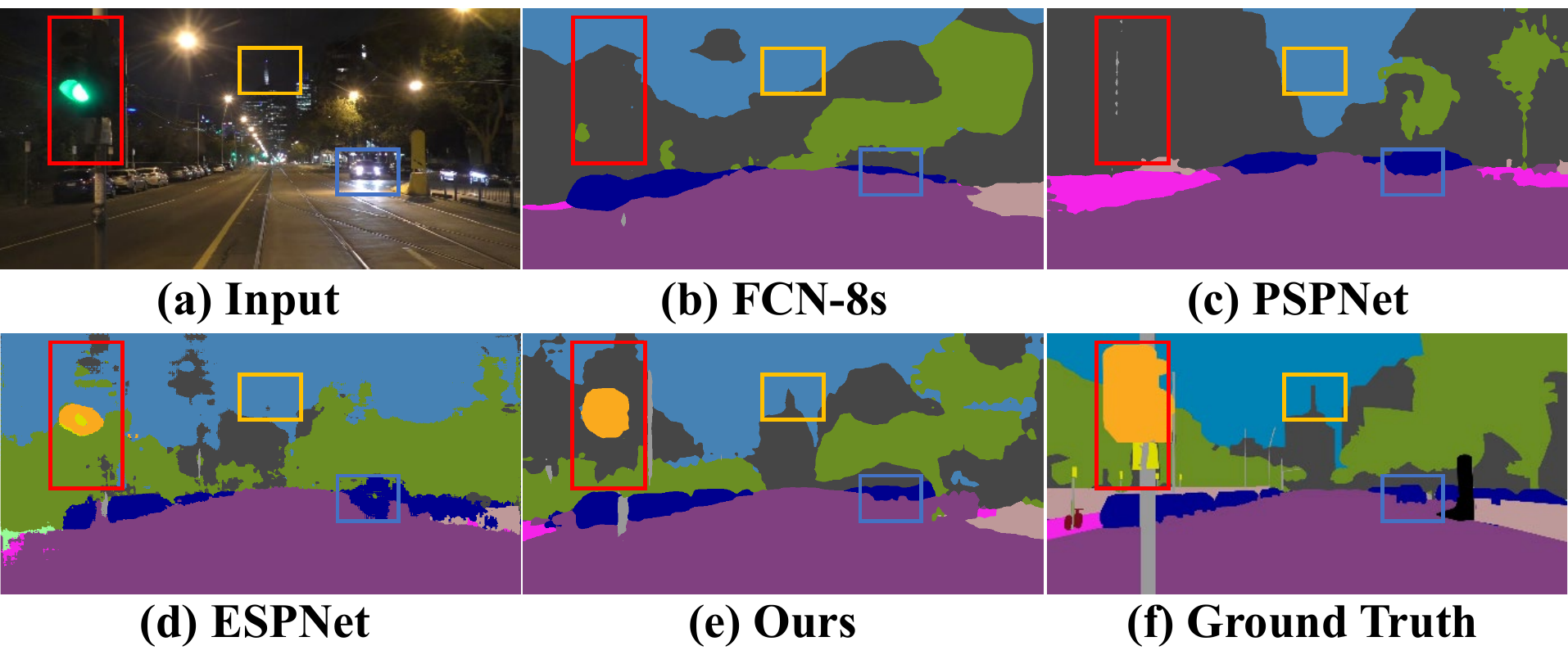}
\caption{Night-time scene analysis. (a) shows an input night-time image, and parsing results from state-of-the-art methods: (b) FCN-8s \cite{Long2015Fully}, (c) PSPNet \cite{zhao2017pspnet}, and (d) ESPNet \cite{Mehta2018ESPNet}. We also show the result by our model (e), and the ground truth (f). The yellow and blue boxes highlight under- and over-exposed regions, respectively. The red box highlights the region with a mixture of under- and over-exposures.}
\label{fig:example}
\vspace{-0.3cm}
\end{figure}

When applied to night-time images, existing scene parsing methods designed for day-time images typically do not perform well.
We observe that night-time scenes often contain both over-/under-exposures,  which can seriously degrade the visual appearances and structures of the input images.
{For} example, Figure \ref{fig:example}(a) shows a night-time scene with both over-exposure (e.g., street lights and car headlights) and under-exposure (e.g., background and regions around the headlights). We can see from Figures \ref{fig:example}(b--d) that state-of-the-art methods are not able to address this problem well.
First, the building highlighted by the yellow box blends into the dark background due to under-exposure, causing it to be difficult to detect.
Second, the texture and structure of the cars highlighted by the blue box are corrupted due to over-exposure, causing them to be difficult to segment correctly.
Third, the traffic light highlighted by the red box is difficult to be detected or segmented correctly due to a mixture of over-/under-exposures.

There are two main challenges to the NTSP problem. First, large-scale \textit{labeled} datasets of \textit{night-time} scenes are not available. Existing large datasets for scene parsing mainly contain day-time images, with few or no night-time images \cite{Cordts2016Cityscapes,Alhaija2017BMVC,brostow2009semantic}.
Models trained on these datasets do not generalize well to the complexity of night-time scenes.
Second, existing methods do not explicitly model over- and under-exposures, as they are primarily developed for day-time scenes. However, as demonstrated in our experiments, explicitly modeling the exposure is necessary in order to achieve robust performances.

To address the first challenge, we propose in this paper a new dataset, named {\it NightCity}, for the NTSP problem. It contains $4,297$
real night-time images of diverse complexity, with pixel-wise annotations. To our knowledge, NightCity is the largest labeled dataset for NTSP, and is an order of magnitude larger than existing datasets for NTSP \cite{GCMA_UAE}. As compared with Cityscapes \cite{Cordts2016Cityscapes}, NightCity covers more diverse and challenging exposure conditions that are typical in night-time scenes. Our experiments show that NightCity can help significantly advance the NTSP performance. It can also serve as a benchmark for evaluating NTSP methods.

To address the second challenge, as we observe that the drop in performance when applying existing scene parsing methods on night-time images is mainly due to the complicated exposure conditions of night-time scenes, we propose an Exposure-Guided Network (EGNet) to explicitly learn over- and under-exposure features to guide the NTSP process. Our model comprises two streams: the segmentation stream and the exposure stream. The segmentation stream learns to predict the semantic label map for the input image, while the exposure stream learns exposure-related features by explicitly predicting the exposure map and uses the learned features to augment the segmentation stream via an attention mechanism.
Experimental results show that our model
outperforms previous methods, achieving state-of-the-art performance on night-time images.

In summary, the main contributions of this paper include:
\begin{itemize}
\item We propose a large-scale labeled dataset of real night-time images for the NTSP problem.
\revisionxin{The dataset will be made publicly available, to help advance research along this direction.}
\item We present an end-to-end exposure-aware NTSP framework, which explicitly learns exposure features to address the NTSP problem.
\item We have conducted extensive evaluations. Our results demonstrate that our proposed model outperforms the state-of-the-art methods on night-time scenes.
\end{itemize}

\section{Related Work}
\subsection{Scene Parsing}
There are a lot of works proposed for scene parsing.
The performance has been improved significantly in recent years due to the development of Convolutional Neural Networks (CNNs).
{For example, in~\cite{wang2017joint}, prior location information at superpixel level was exploited with CNNs to strengthen object discriminativity in scene parsing.}
{Multi-level-based methods \cite{Long2015Fully,zhang2018exfuse,lin2017refinenet,lin2018multi,cheng2019spgnet} were widely used by learning multi-level features to extract the global context and to preserve the low-level details.}
Recently,
attention-based methods \cite{fu2018dual, takikawa2019gated, li2019expectation} have shown promising performances.
Fu et al. \cite{fu2018dual} adaptively integrated local features with their global dependencies using position attention and channel attention modules.
{
Huang et al. \cite{huang2018ccnet} used a novel criss-cross attention module to model long-range contextual dependencies over local feature representations. {
Choi et al. \cite{choi2020cars} adopted the height attention to learn different categories that are distributed at different heights.}
Two-stream approaches were also proposed \cite{Pohlen2016Full,takikawa2019gated}. For example, 
Takikawa et al. \cite{takikawa2019gated} introduced a two-stream network, with one of the streams explicitly wiring} shape information for scene parsing.
{
Feng et al. \cite{9444191} transferred the knowledge of a teacher stream to a student stream via a pixel-wise
similarity distillation module and a category-wise similarity distillation module, for balancing the accuracy and inference time.}

Many methods are proposed to learn robust object representations for scene parsing.
%
Yuan et al.\cite{yuan2019object} exploited object-contextual representations by exploring the features of corresponding object classes, to characterize a pixel. Zhu et al. \cite{Zhu_2019_CVPR} exploited video prediction models to predict the class labels.
{GPSNet \cite{9318517} was designed to adaptively select receptive fields while maintaining a dense sampling capability for scene parsing.
%
Huang et al.~\cite{Huang2020} proposed a Scale-Adaptive Network to tackle the varying scale problem of objects.
To bridge the gaps between the domain of (limited) training data and that of test scenes, some domain adaptation-based methods \cite{zhang2017curriculum, wang2019weakly, liu2021source, barbato2021latent} were proposed to utilize useful knowledge of synthetic scenes to enrich the scene representations.}

All the above methods achieved state-of-the-art results on day-time datasets. However, unlike these existing works, we focus our attention on night-time scenes with poor lighting conditions in this work.

\subsection{Scene Parsing in Adverse Conditions}
Although most existing works focus on the ``normal'' scenarios with well-illuminated scenes, there are also some works that address the challenging scenarios.
For rainy night scenes, \cite{di2020rainy} tried to solve the rainy night scene parsing problem via transferring day-time knowledge. It collected $226$ images with eight categories. {
Tung et al. \cite{tung2017raincouver} also tried to handle this problem and proposed 95 annotated night-time images with three categories.}
{
Zheng et al. \cite{zheng2020forkgan} presented a Fork-shaped Cyclic generative module, decoupling domain-invariant content and domain-specific style, to translate the scenes in adverse condition to images with better visualizations for parsing. }
For multiple and mixed adverse conditions, Valada et al. \cite{valada2017adapnet} proposed the convoluted mixture of deep experts fusion techniques to understand these adverse conditions, including rain, snow, sunset and night scenes among 13 categories. {WildDash \cite{zendel2018wilddash} and BDD100K \cite{BDD100K} aimed to test segmentation performances by presenting many kinds of scenarios with 13 and 345 night-time images, respectively.}
Sakaridis et al. \cite{GCMA_UAE,sakaridis2020map} proposed guided curriculum model adaptation to solve the night-time semantic segmentation problem, with
a small dataset of 151 night-time images.
{Recently, 
Wu et al. \cite{wu2021dannet} proposed an unsupervised one-stage adaptation method for NTSP, by exploiting adversarial learning between labeled day-time dataset and unlabeled dataset of day-night aligned image pairs.}

In contrast to the above works, we propose a large-scale dataset of real night-time images with semantic annotations {($21$ categories)} and a novel exposure-aware framework to address the NTSP problem.

\begin{figure*}[t]
\includegraphics[width = \linewidth]{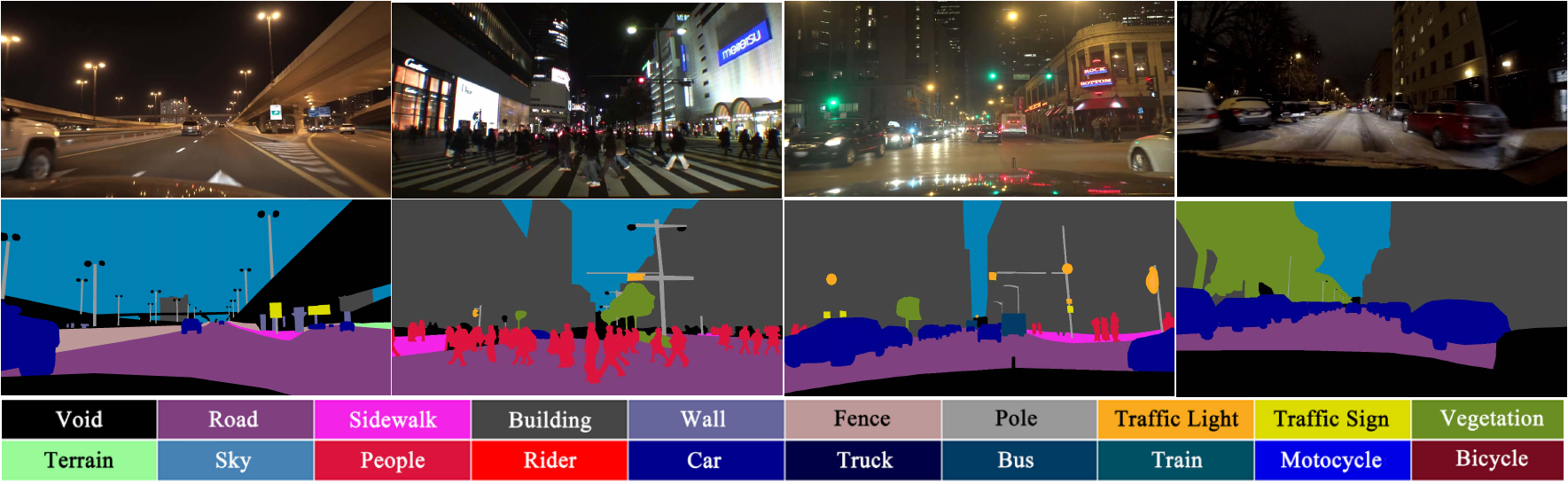}
\caption{Several example images from our NightCity {dataset}. Note that some regions around the car headlights are over-exposed and some background regions are under-exposed.}
\vspace{-3mm}
\label{fig:darkcity}
\end{figure*}

\subsection{Image {Enhancement/Correction}}
One naive solution to our problem is to first apply image enhancement on the input night-time images and then perform scene parsing with an existing day-time method. Image enhancement methods, e.g., \cite{yan2016automatic,Wang2013Naturalness,Li2017LightenNet,Ying_2017_ICCV}, aim to remap the pixel values to improve the image visibility.
Wang et al. \cite{Wang_2019_CVPR} tried to address the under-exposure problem by estimating an illumination map.
Cai et al. \cite{Cai2018Learning} proposed to learn a deep image contrast enhancer from multi-exposure images.
In \cite{Yang2018Image}, an end-to-end network was proposed to convert an input LDR image first to HDR to recover the missing details due to under-/over-exposure, and then reproject it back to LDR as output while preserving the recovered details. {Xu et al. \cite{xu2020learning} proposed a frequency decomposition framework to address the practical under-exposure problem with noise.}
However, night-time images often contain both under-/over-exposed regions (with pixel values very close to zero/one). The remapping process may not recover meaningful values.
As demonstrated in our experiments in Section~\ref{sec:beni_night_data}, pre-processing the input images with a state-of-the-art image enhancement method before scene parsing cannot address our problem well.

{In contrast, our method does not apply any pre-processing techniques. It achieves the state-of-the-art NTSP performances by (1) constructing the first large-scale NTSP dataset, and (2) exploiting exposure-aware features to enrich night-time scene representations via the proposed network.}

\section{The NightCity Dataset} \label{sec:nightcity}
To construct our dataset, we first collect real night-time driving videos {(which were captured using a Driving Recorder during car driving) over the Internet from various cities, e.g.,  Los Angeles, New York, Chicago, Hong Kong, London, Tokyo and Toronto}. These videos cover urban street, highway and tunnel scenarios. We then manually select $4,297$ diverse images with no obvious motion blur from these videos for manual annotation, following the approach used to construct the Cityscapes dataset \cite{Cordts2016Cityscapes}. \revisionxin{Table \ref{table:city_dis} shows the distribution of the videos over the considered cities.}

\renewcommand{\arraystretch}{1.2}
\begin{table}[h]
\begin{center}
\caption{\revisionxin{Distribution of the videos over all considered cities.} }\label{table:city_dis}
\begin{small}
{
\begin{tabular}{l c c c c}
\hline 		
  City & Number of images &Train &Test   \\
\hline
Los Angeles  &148 &136 &12 \\
\hline
New York &375 &261 &114 \\
\hline
Chicago &195 &136 &59\\
\hline
Toronto &132 &92 &40 \\
\hline
Melbourne &183 &128 &55\\
\hline
London  &74 &52 &22\\
\hline
Dubai &512 &357 &155\\
\hline
Helsinki &659 &460 &199\\
\hline
Hong Kong &367 &256 &111\\
\hline
Seoul &344 &240 &104 \\
\hline
Nagoya &1025 &715 &310\\
\hline
Tokyo &181 &126 &55\\
\hline
Other cities in Japan &102 &72 &30 \\
\hline
\end{tabular}}
\end{small}
\end{center}
\vspace{-2.5mm}
\end{table}

\textbf{Annotation}. Like Cityscapes, we annotate semantic regions as layered polygons using LabelMe \cite{russell2008labelme}. All our images are of resolution 1024$\times$512. Given the {difficulty} of recognizing objects in the over-/under-exposed regions, we annotate our images by two separate annotators (A and B) and re-evaluate their results by a third one (C). Annotators may refer to the corresponding video if an image is difficult to see.
For each image $I \in R^{H \times W \times 3}$,
annotators A and B give the annotations $G_A$ $\in R^{H \times W \times C}$ and $G_B$ $\in R^{H \times W \times C}$, respectively.
Annotator C {compares} the difference between ${G_A}$ and ${G_B}$ to produce a compromised map.
{There are two {situations} where Annotator C may correct a label. First, the two labels by the two annotators are different. Second, the two labels by the two annotators are the same, but Annotator C considers these labels as incorrect. In this case, C would discuss with the two annotators to come up with the final label.}

\revisionxin{The steps for Annotator C to follow are summarized as:
\begin{enumerate}
\item If C considers both $G_A$ and $G_B$ are correct, there will be no changes.
\item If $G_A$ and $G_B$ are different and C considers one is correct and the other is incorrect, C will simply select one of them as the final annotation (i.e., majority win).
\item If $G_A$ and $G_B$ are different but C considers both incorrect. C will discuss with Annotators A and B to make the final decision.
\end{enumerate}
}

\begin{figure*}
\centering
\includegraphics[width = 0.7\linewidth, height=5cm]{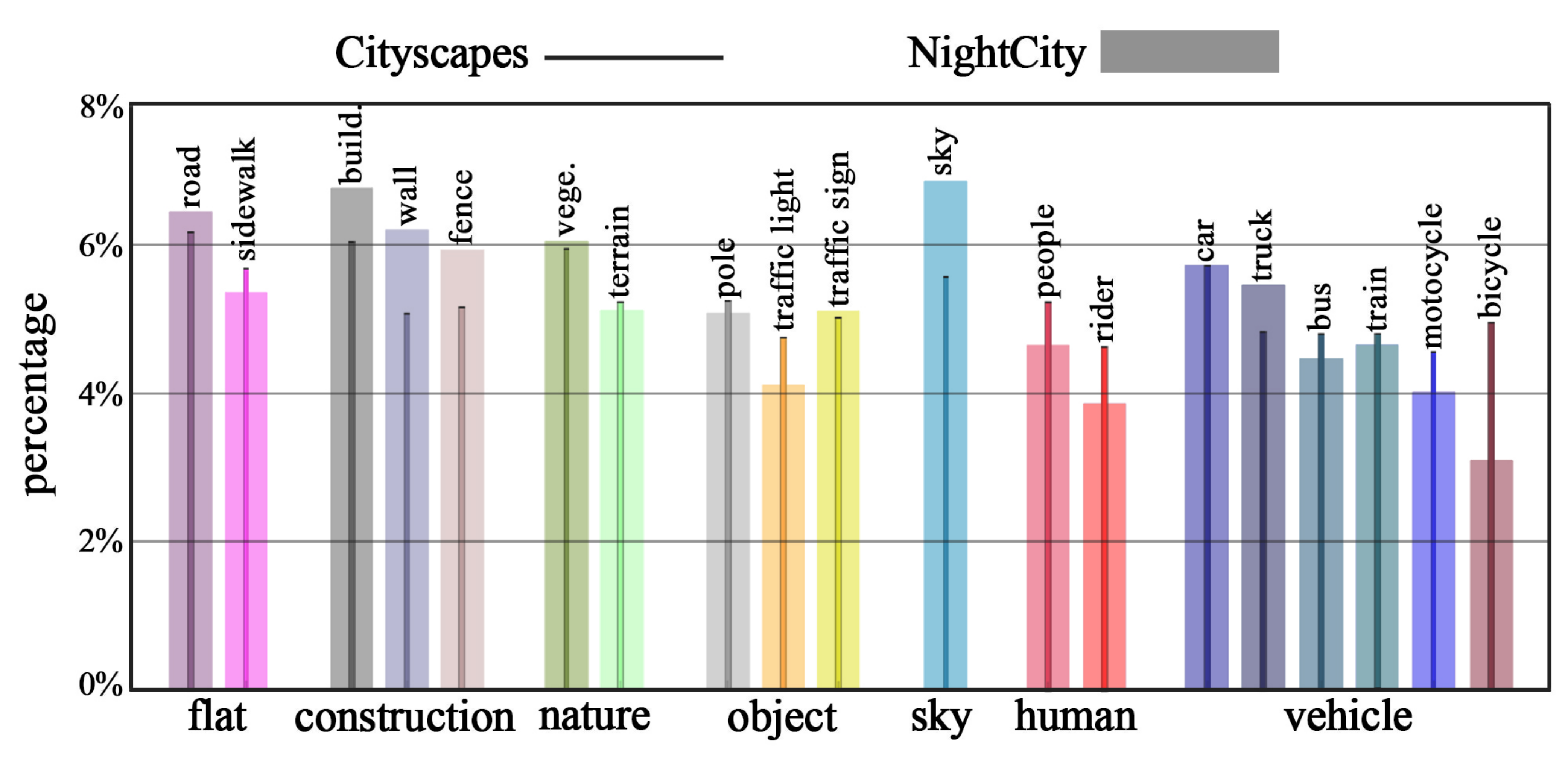}
\caption{Distributions of labeled pixels for each semantic class in Cityscapes and NightCity\revisionxin{, shown in log scale.}
}
\label{fig:distribution}
\vspace{-0.2cm}
\end{figure*}

{Statistically, there are in total 9.8\% of the pixels that are labeled differently by the two annotators. Among these different annotations, 47.9\% are finally corrected, i.e., (total number of different-labeled pixels finally corrected)/(total number of different-labeled pixels).
The error proportions, i.e., (total number of error pixels per class)/(total number of pixels per class), of five object categories that are most difficult to annotate are shown in Table \ref{table:error_pro}. We find that poles have the highest annotation error proportion, as they can be easily ignored due to their small size. In addition, buildings and sky can be easily mistakable with each other. }

\renewcommand{\arraystretch}{1.2}
\begin{table}[tb]
\begin{center}
\caption{\revisionxin{Proportions of annotation errors of five most difficult object categories to annotate. }}\label{table:error_pro}
\begin{small}
{
\begin{tabular}{l c c c c c}
\hline 		
  Class &Pole & Building &  Wall  & Sky & Terrain  \\
\hline
Error(\%) & 6.38 &4.76 &4.84 & 4.44 &3.19 \\
\hline
\end{tabular}}
\end{small}
\end{center}
\vspace{-2.5mm}
\end{table}

Some regions that are too difficult to annotate even by humans {are} labeled as \textit{invalid regions} so that they are ignored during training and evaluation.  In GCMA \cite{GCMA_UAE}, a pixel is labeled as invalid if the annotator considers it as invalid. In our paper, we decide it based on the judgement of multiple annotators, i.e., if at least two of the three annotators consider a pixel as invalid (a majority-win strategy). The total proportion of invalid pixels in NightCity is 7.36\%.

Figure \ref{fig:darkcity} shows some example images from NightCity, demonstrating that our images are from complex scenes and their contents are difficult to recognize and segment even for humans, due to the under-/over-exposure problems caused by insufficient lighting, street lights and car headlights. Hence, our NightCity dataset represents a rather challenging training and evaluation dataset for NTSP.

\textbf{Object Class Distribution.} Figure \ref{fig:distribution}
compares the distributions of labeled \revisionxin{pixels in NightCity and Cityscapes. As some categories have significantly higher numbers of pixels (e.g., about $1\times 10^9$ pixels belonging to``building'') than some others (e.g., about $1\times 10^7$ pixels belonging to``motorcycle''), we show the distributions in log scale.}
Overall, the pixel distributions of all classes in our night-time street scenes are similar to those of the day-time street scenes, except for Bicycle. This is reasonable because there are typically fewer bicycles on the street at night.

\begin{figure}[t]
\begin{center}
\includegraphics[scale=0.36]{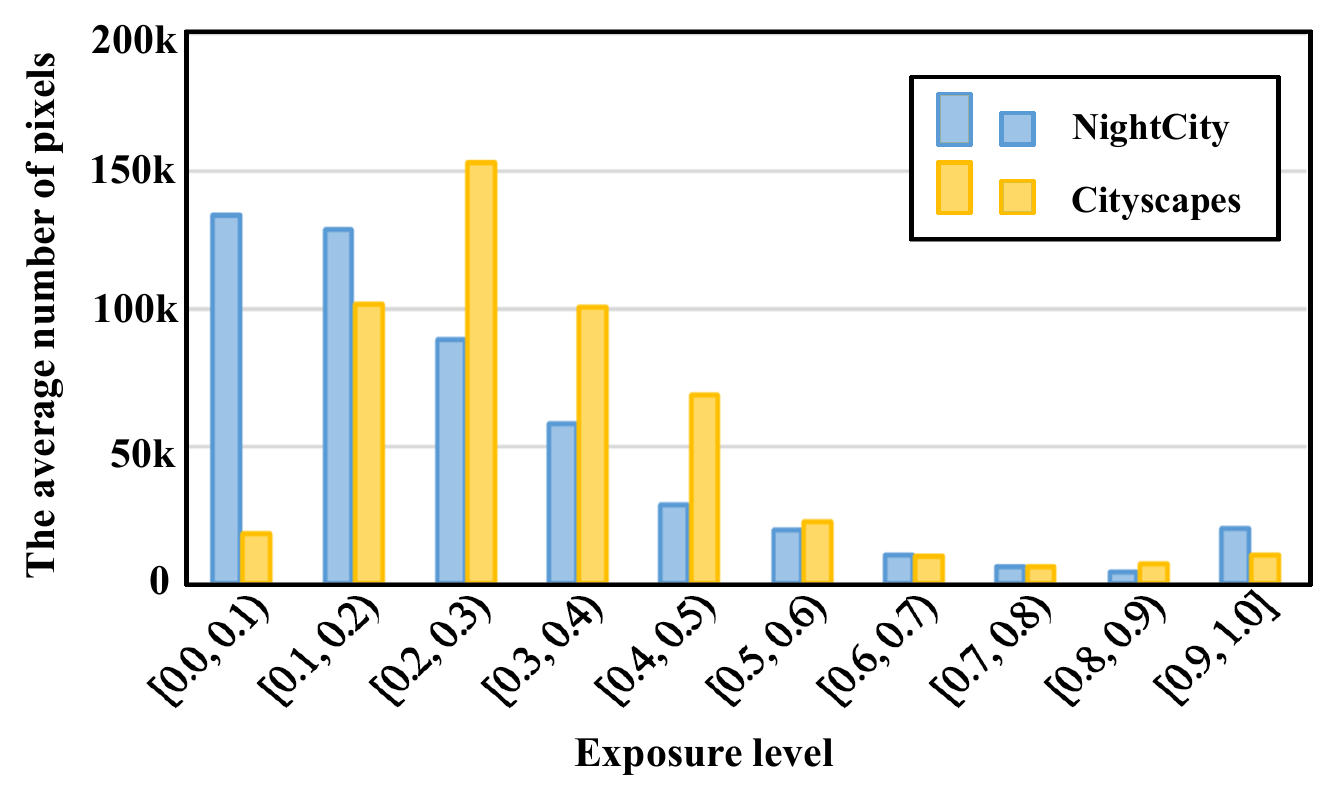}
\caption{The {average} number of pixels per image at each exposure level for Cityscapes and for NightCity, {with image resolution of 1024 $\times$ 512}.}
\label{fig:exposure_infor}
\end{center}
\vspace{-0.5cm}
\end{figure}

\textbf{Exposure Distribution.} To reveal the over-/under-exposure problems in our dataset, we also analyze the exposure conditions in our images by checking the pixel values.
In photography, the exposure is determined by the shutter speed, lens aperture, scene luminance and ISO number.
However, as this information is unknown to us, we use the V channel (i.e., intensity) of the images in the HSV color space to represent the exposure. In particular, we divide the exposure value equally into ten bins from 0 to 1 with an interval of $0.1$. Figure \ref{fig:exposure_infor} shows the {average} number of pixels {per image} that falls into each bin for NightCity and for Cityscapes. The [0, 0.1] bin stores the most under-exposed pixels, while the [0.9, 1.0] bin stores the most over-exposed pixels. From Figure \ref{fig:exposure_infor}, we can see that NightCity has significantly more under-exposed pixels than Cityscapes. Meanwhile, we observe that Cityscapes has a moderate exposure condition, with most of pixels falling between [0.2, 0.6]. Although the two datasets have a similar number of over-exposed pixels, we notice that the over-exposed regions in Cityscapes are mostly in the sky, which is easy to predict, while the over-exposed regions in NightCity can be produced by the street lights, car headlights, or traffic lights, which are very difficult to differentiate. This indicates that the NightCity dataset has various challenging exposure conditions, compared with Cityscapes.

\begin{figure*}[t]
\centering
\includegraphics[width=1.0\linewidth]{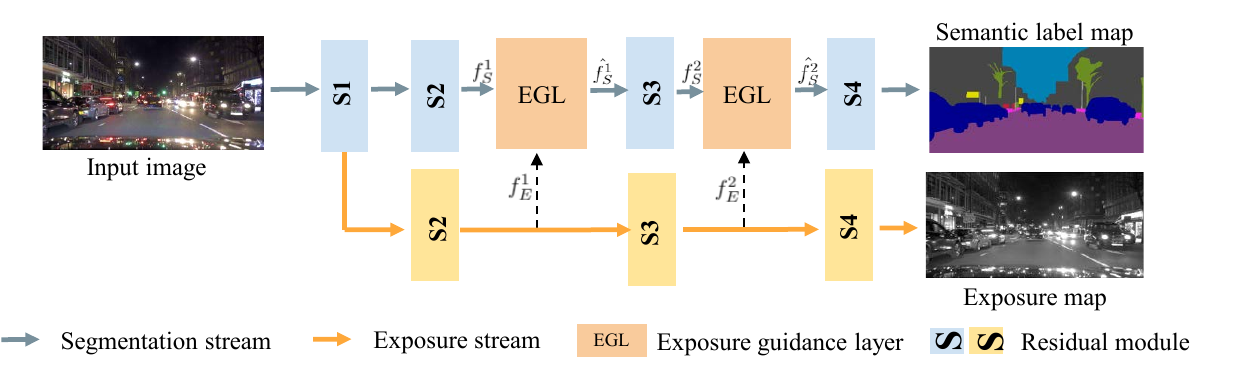}
\caption{The architecture of our exposure-guided network. It contains two streams, a segmentation stream in blue to predict a semantic label map, and an exposure stream in yellow to predict an exposure map. {We introduce the exposure guidance layer (EGL) to augment segmentation features with exposure features.}}
\label{fig:flow}
\end{figure*}

\textbf{Size and Splitting.} Compared with existing datasets under adverse conditions, the number of images in NightCity is considerably higher. For example, \textit{Foggy Driving} \cite{SDV18} proposes a real fog dataset with a total of 101 foggy images for testing (of which only $33$ images are finely annotated and the remaining 68 images are only coarsely annotated).
\textit{The Dark Zurich Dataset} \cite{GCMA_UAE}  contains only $151$ night-time images with pixel-level annotations.
\revisionxin{BDD100K \cite{BDD100K} contains $345$ night-time images with some labeling errors. Hence, their ground truth is not very reliable.}
\revisionxin{\textit{Raincouver} \cite{tung2017raincouver} contains 95 coarsely annotated night-time images but only with 3 classes.}
\revisionxin{\textit{WildDash} \cite{zendel2018wilddash} only has 13 finely annotated night-time images.}
Instead, all $4,297$ images in our NightCity dataset are manually annotated with fine class labels. Thus, NightCity is the largest dataset of real night-time images with high-quality pixel-level annotations.

The NightCity dataset is split into training and test sets. We split them
in such a way that they preserve similar distributions to the whole dataset. In this way, our training and test splits include $2,998$ and $1,299$ images, respectively.

\section{Exposure-Guided Network}\label{sec:network}
The core idea of our proposed Exposure-Guided Network (EGNet) is to explicitly learn exposure features and use them to augment the segmentation process. Hence, our network is designed to have two coupled streams: exposure stream and segmentation stream. The exposure stream learns to predict where exposure occurs and {uses} the predictions to guide the segmentation stream via the exposure guidance layers so that the segmentation stream can discriminate the under-/over-exposed regions more effectively.

\def\imh{.9in}
\begin{figure*}
\renewcommand{\tabcolsep}{1pt}
\begin{center}
    \begin{tabular}{cccc}
    \includegraphics[width = 0.245\linewidth, height=\imh]{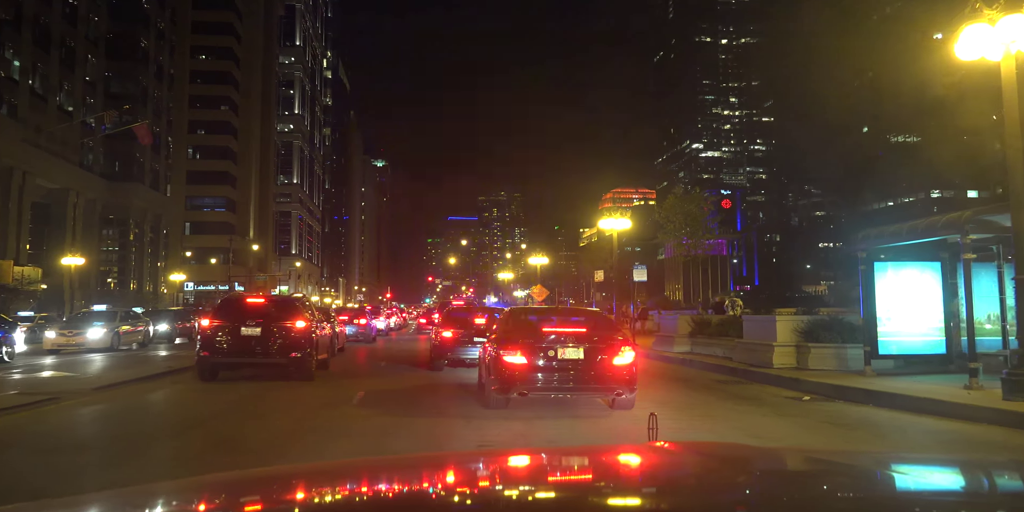}&
    \includegraphics[width = 0.245\linewidth, height=\imh]{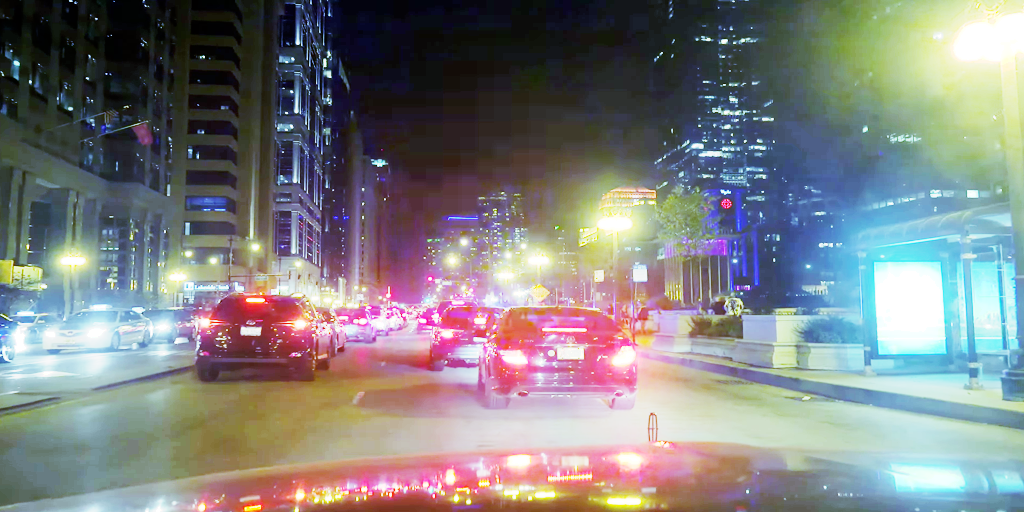}&
    \includegraphics[width = 0.245\linewidth, height=\imh]{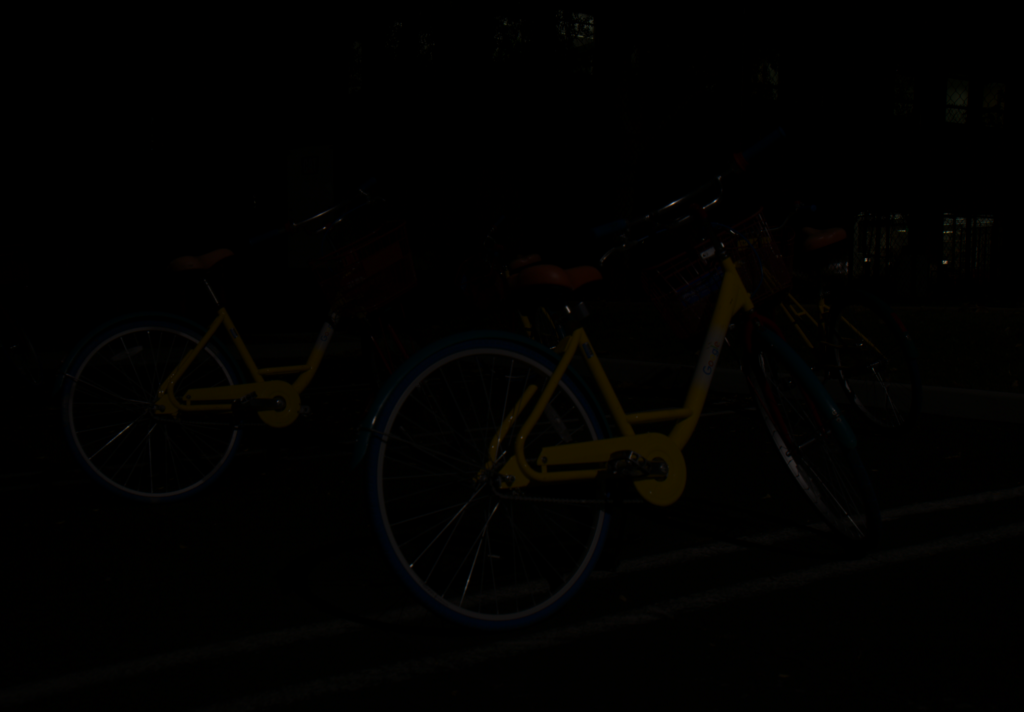}&
    \includegraphics[width = 0.245\linewidth, height=\imh]{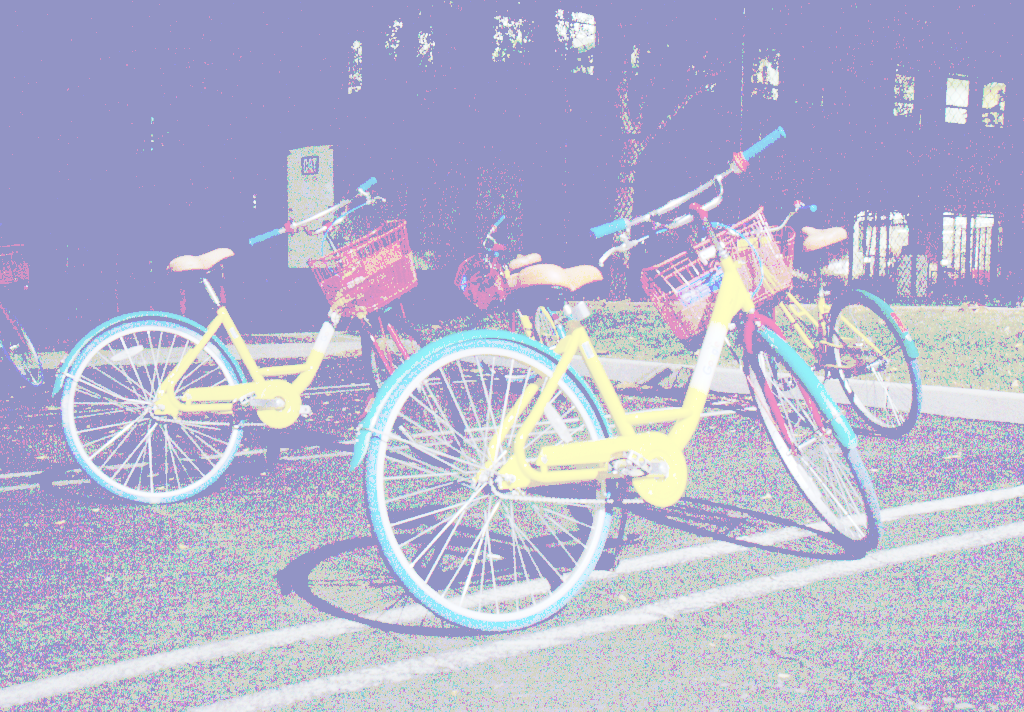}\\
    \includegraphics[width = 0.245\linewidth, height=\imh]{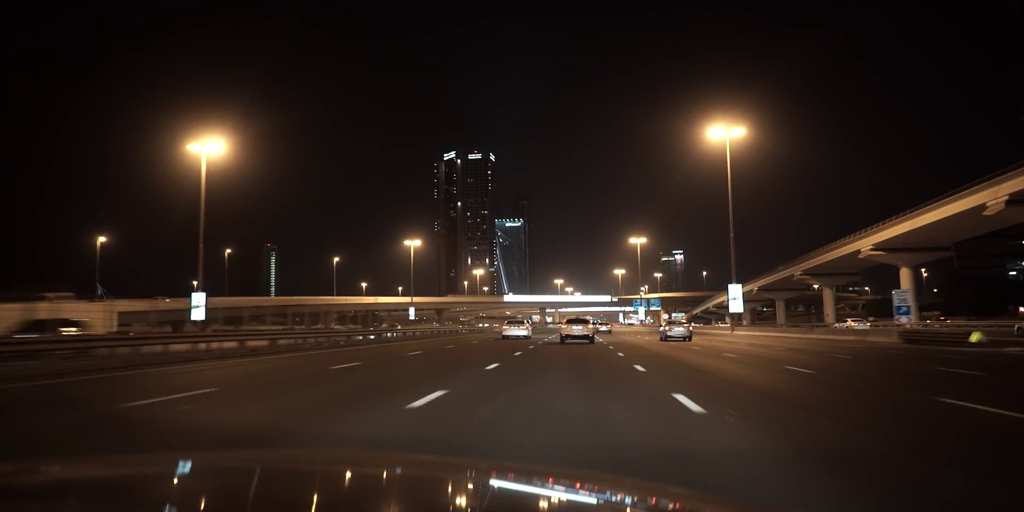}&
    \includegraphics[width = 0.245\linewidth, height=\imh]{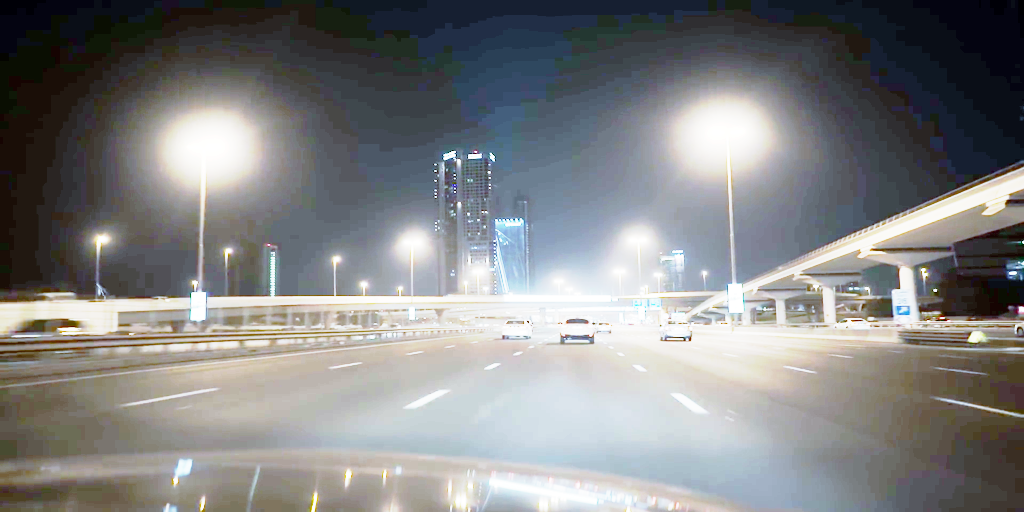}&
    \includegraphics[width = 0.245\linewidth, height=\imh]{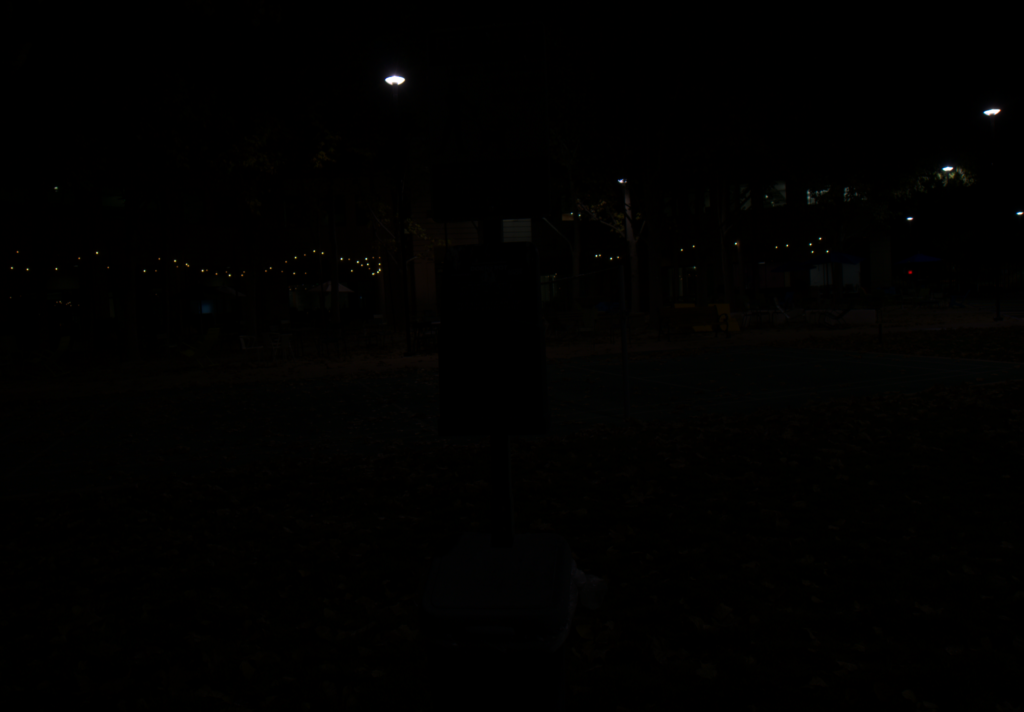}&
    \includegraphics[width = 0.245\linewidth, height=\imh]{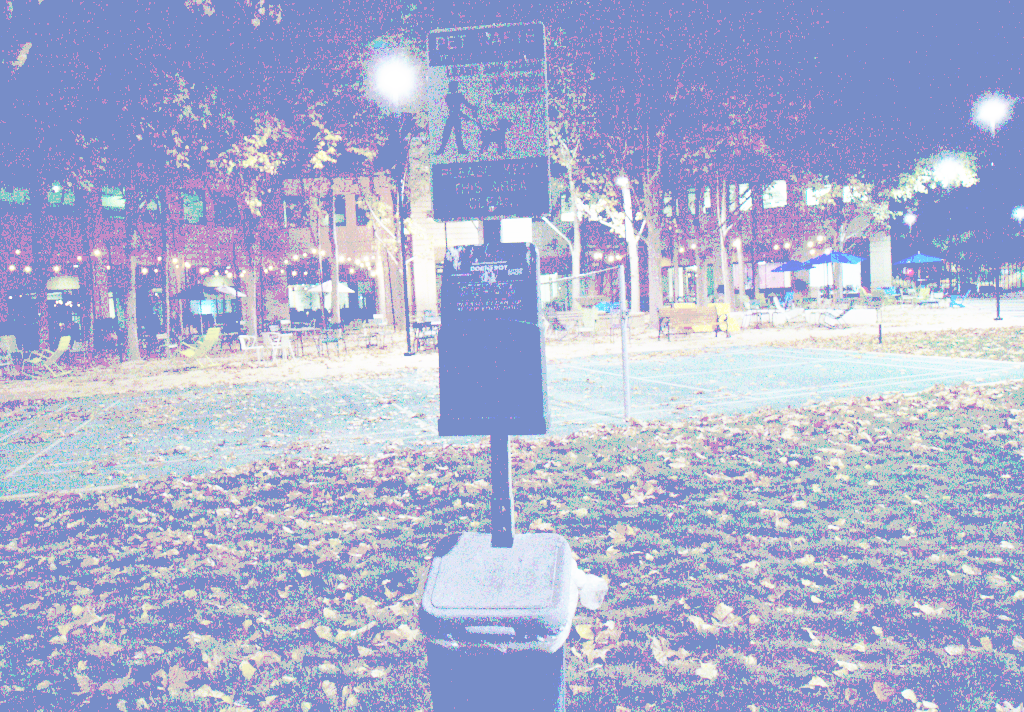}\\
    \includegraphics[width = 0.245\linewidth, height=\imh]{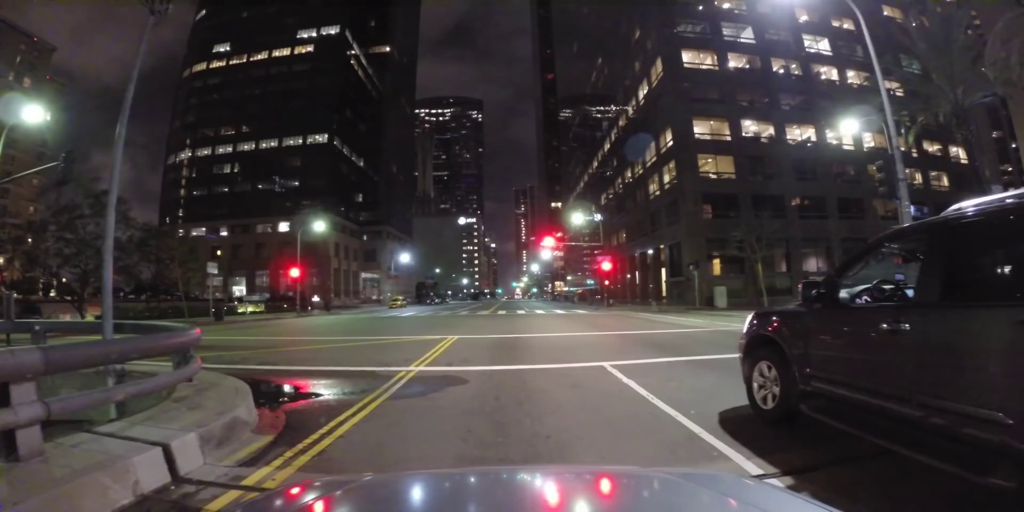}&
    \includegraphics[width = 0.245\linewidth, height=\imh]{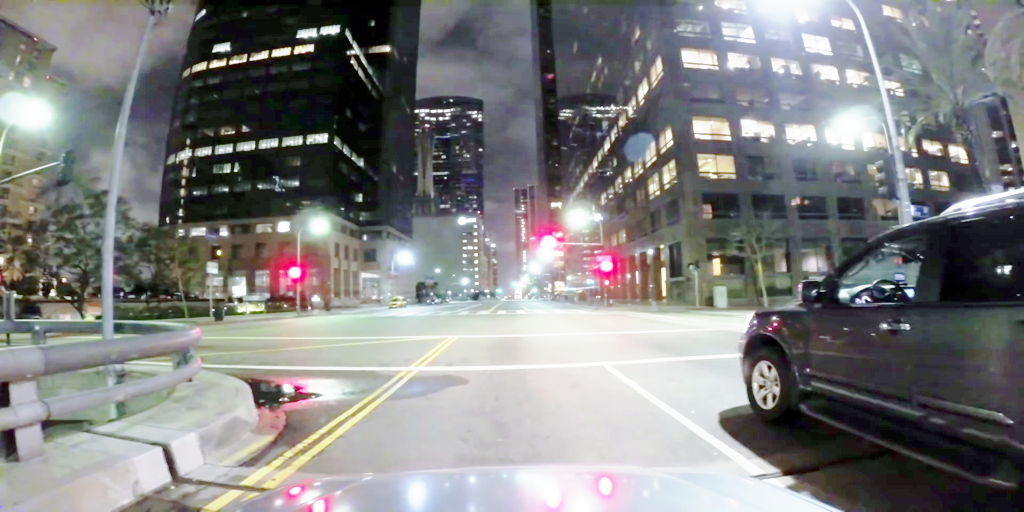}&
    \includegraphics[width = 0.245\linewidth, height=\imh]{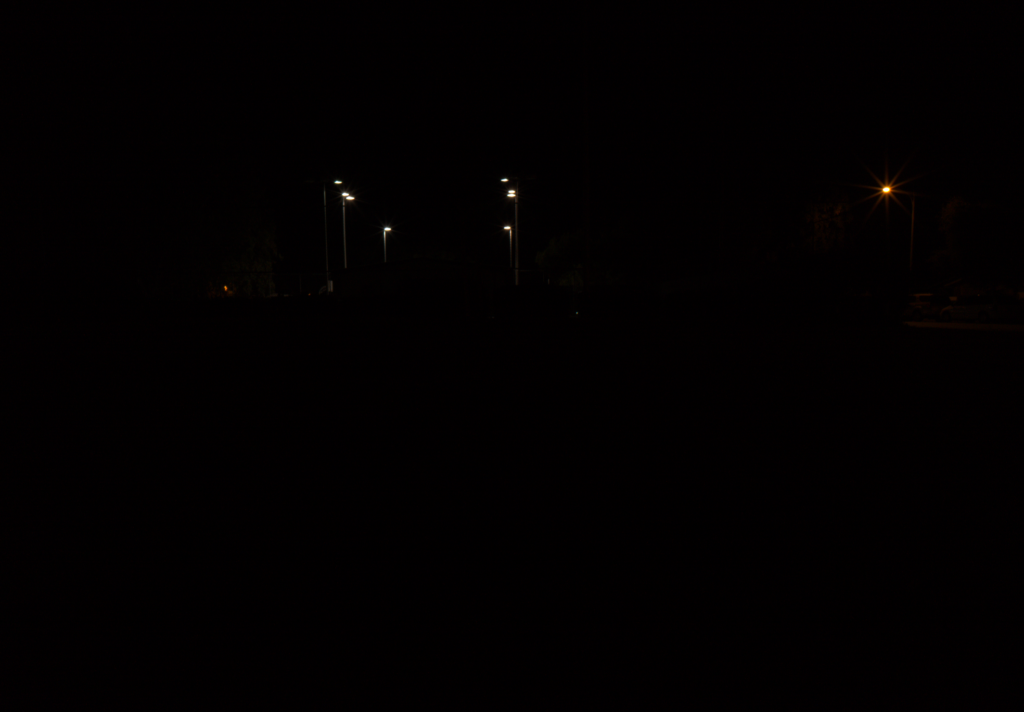}&
    \includegraphics[width = 0.245\linewidth, height=\imh]{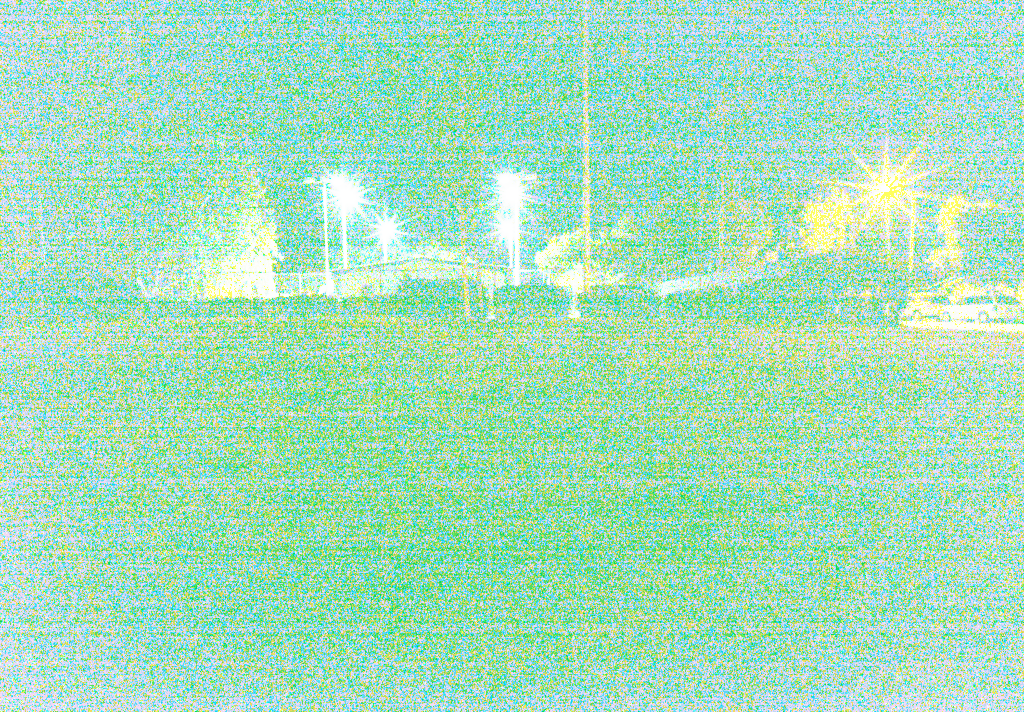}\\
    (a) Night-time Scenes& (b) Hist.Eq. of (a) & (c) Low-illuminated Scenes & (d) Hist.Eq. of (c)\\
    \end{tabular}
\end{center}
\caption{Comparison of our night-time urban scenes (a) and low-illuminated scenes (c) from~\cite{xu2020learning}, and their enhanced ones using histogram equalization (see (b) and (d), respectively). Noise is negligible in the night-time urban scenes but non-negligible in the low-illuminated scenes.}%
\vspace{-3mm}
\label{fig:nighttime_lowlingt}
\end{figure*}

\subsection{Network Architecture}
Figure \ref{fig:flow} shows our network architecture. Given an input image $I \in R^{H \times W \times 3}$, we use a backbone encoder to transform it into convolutional features. The network is then split into the segmentation stream (top) and exposure stream (bottom). The segmentation stream predicts a class label map $M_S \in R^{H \times W \times C}$, where $C$ is the number of classes. The exposure stream also outputs a pixel-wise exposure map $M_E \in [0, 1]^{H \times W}$, indicating the magnitude of exposure at each pixel.

As discussed in Section \ref{sec:nightcity}, we approximate the exposure map using the normalized V channel of the input image in the HSV color space.

Our network is mainly based on ResNet \cite{He2015}, like most existing scene parsing models~\cite{CP2016Deeplab,Peng2017Large,Yu2018Learning,zhao2017pspnet}. ResNet has four stages that extract hierarchical features at different scales, with earlier stages capturing low-level features and later stages capturing high-level semantics. In particular, we take stage1 (S1) of ResNet-101 as our backbone encoder. The segmentation and exposure streams have the same architecture, by combining stage2 (S2), stage3 (S3) and stage4 (S4) of ResNet, except for the last output layers. For the output layer, the segmentation stream uses a 19-channel convolutional layer  to output a semantic label map, while the exposure stream uses another sigmoid nonlinearity function to output a soft binary exposure map.
{To train the network, we use a cross-entropy loss for the segmentation stream, and a $\ell_1$ loss for the exposure stream. Our final loss is defined as}
$L = {\alpha}{L_c}+{\beta}{L_e}${, where $\alpha$ and $\beta$ are weights of the two losses. }

\subsection{Exposure Guidance Layer}
To extract exposure features from the exposure stream to guide the segmentation stream, we introduce the exposure guidance layer (EGL) to augment the intermediate features of the {segmentation} stream.
Let $f_S$ and $f_E$ be the {intermediate features} of the segmentation and exposure streams.
EGL updates $f_S$ to obtain {exposure-aware} features $\hat{f_S}$ as:
\begin{equation}
\hat{f_S} = {w_1}{f_S} + {w_2}{f_S} \otimes {W_r},
\label{eq:attention}
\end{equation}
where $\otimes$ {denotes} element-wise multiplication. $w_1$ and $w_2$ are weight parameters.
${W_r}$ = ${\delta}({W*f_E}+b)$ is a soft spatial attention map as in the popular works \cite{Chen2016Attention,Chu2017Multi}, where $W$ and $b$ are learnable parameters,  and $\delta(\cdot)$ is a sigmoid function. The exposure-aware features $\hat{f_S}$ are then {fed into the next stage of the} network.
As shown in Figure \ref{fig:flow}, both $f_S$ and $f_E$ are from the previous stages.
All {the} operations in EGL are differentiable so that we can train the network end to end. In addition, {EGL enables the gradients to be back-propagated from the output exposure map to the segmentation stream, thereby allowing the segmentation stream to exploit exposure information.}

The intuition behind Eq.~\ref{eq:attention} is that our model needs to learn how to weight the features in the segmentation stream based on the exposure features, in order to generate more discriminative segmentation features particularly at under-/over-exposed regions. This formulation also forces our model to learn the exposure features that help predict the exposure maps and guide the scene parsing task towards an optimal performance.

\subsection{{Discussion: Is Noise Always Vital In Night-time Scenes?}}
{Noise is inevitable in every imaging pipeline. The intuition is that noise is fatal in night-time scenes due to a lack of illumination, in which cases users may have to increase the ISO value to capture more photons.}
{However, we find that in our night-time urban scenes, the noise issue could be ignored. This is because in night-time urban scenes, there are typically some man-made light sources, such as streetlights and headlights, appearing in the scene. The illuminance of the camera in this type of scenes is typically larger than 10 Lux. With a sufficient number of photons overall in the scene, noise could be ignored even at a low ISO value due to a relatively high signal-to-noise ratio.}

{For illustration, a group of visual comparisons between our night-time urban scenes and low-illuminated scenes from~\cite{xu2020learning} are shown in Figure~\ref{fig:nighttime_lowlingt}. We can see that due to the existences of man-made light sources, the illuminance in an auto-exposed camera is typically larger than 10 Lux, resulting in a negligible noise level (both in the original and histogram equalized night-time images).}
{In contrast, noise can be a significant problem in the low-illuminated scenes if there is an insufficient number of photons captured by the camera due to a short exposure duration or a lack of light sources in the scene. In this situation (where the illuminance at the camera is typically less than 1 Lux), noise cannot be ignored due to the low signal-to-noise ratio.}
{Hence, our method does not require any denoising mechanisms for the night-time scene parsing task.}

\section{Experiments}\label{experiments}
{In this section, we first introduce the exposure-aware Fi-score (EF1) as one of the evaluation metrics.
We then compare our method to the mainstream day-time scene parsing methods, to reveal their limitations when applied to the night-time scene parsing task.
Next, We justify the necessity of the proposed NightCity dataset.
We further verify the effectiveness of the proposed model by comparing our method to ad hoc night-time scene parsing methods.
Finally, we provide ablation studies to analyze different components of our proposed model.}

{\bf Implementation Details.} Our model is trained using the SGD optimizer with a batch size of $6$ and an initial learning rate of 1e-5. We decrease the learning rate using a polynomial policy with a power of $0.9$. Our model is trained for $40,000$ iterations, which takes about $12$ hours on a PC with an i7-7700K CPU and two Nvidia 1080Ti GPUs. We set the image resolution to $500\times500$ for training and $900\times900$ for testing.
All predictions are {scaled} to $1024\times512$, same resolution as the original image.
We set $\alpha$, $\beta$, $w_1$ and $w_2$ to 1, 0.01, 1, 0.3, respectively.

\begin{table*}
\centering
\caption{Comparison of our model with state-of-the-art {Day-time} scene parsing methods on the NightCity test set. ``C'' Columns: trained on Cityscapes. ``C (IE)'' columns: trained on Cityscapes, and applying DRHT~\cite{Yang2018Image} as an image enhancement (IE) pre-process during testing. ``N'' columns: trained on NightCity. ``C + N'' columns: trained on both Cityscapes and NightCity. The best results are marked in \textbf{bold}.}
\label{table:result_tabel1}
\resizebox{0.65\linewidth}{!}{
\begin{tabular}{l|*{7}{c|}c}
\hline
\multirow{3}*{Methods} &\multicolumn{4}{c}{mIoU (\%) $\uparrow$} &\multicolumn{4}{|c}{mEF1  $\uparrow$}\\
\cline{2-9}& & & & & & & & \\[-11pt]
&{C}&{C (IE)}&{N}&{C + N}&{C}&{C (IE)}&{N}&{C + N}\\
\hline
SegNet \cite{Badrinarayanan2017SegNet}&  5.7  & 5.6   &17.3  &18.1   &0.38  & 0.41  & 0.67 &0.68  \\
FCN-8s \cite{Long2015Fully}&8.1    & 8.2   &28.2  & 28.1  &0.58  &0.58   &0.80  &0.79\\
PSPNet \cite{zhao2017pspnet}& 12.6   & 11.1   & 46.3 &  46.5 & 0.61 & 0.54  & 0.87 & 0.86\\
BiSeNet \cite{yu2018bisenet}& 6.1   &  8.8  & 50.0  & 46.2  & 0.42  &  0.53 & 0.87 & 0.85\\
PSANet \cite{zhao2018psanet}&  7.8  &8.2    &28.8  & 28.5  &0.53  &0.57   &0.70  &0.70\\
ESPNet \cite{Mehta2018ESPNet}&9.1    &8.9    &34.2  &34.5   &0.28 &0.28   &0.82  &0.83\\
DFN \cite{Yu2018Learning}& 9.6   &  9.6  & 50.1 & 52.3  & 0.33 & 0.34  &0.87  &\textbf{0.88}\\
CCNet \cite{huang2018ccnet}&11.1 &10.9&49.2&49.8&0.42&0.46&0.85&0.86 \\
\revisionxin{VPLR} \cite{Zhu_2019_CVPR} &12.5 &11.2&50.4&52.4&0.65&0.60&0.87&\textbf{0.88} \\
\revisionxin{HANet} \cite{choi2020cars}  &13.4 &11.8&51.1&53.1&0.64&0.59&0.87&0.86 \\
\hline
Ours &-    & -   &\textbf{51.8}  & \textbf{53.9}  & - &  - &\textbf{0.88}  &\textbf{0.88}   \\
\hline
\end{tabular}}
\vspace{0.05in}
\vspace{-0.001in}
\end{table*}

\subsection{Evaluation Metrics}\label{sec_eva_metr}
Following previous works \cite{Long2015Fully,CP2016Deeplab,Peng2017Large,lin2017refinenet}, we use mean IoU (mIoU) to evaluate the NTSP performance in our experiments. In addition, we propose an exposure-aware F1-score (EF1) to evaluate the segmentation performance under different exposure levels. EF1 takes both recall and precision into consideration. In particular, as in Figure \ref{fig:exposure_infor}, we divide night-time images into $G$ groups according to the exposure levels. Hence, EF1 for group $g$ is formulated as:
\begin{equation}
\label{eq:metric_f1g}
    EF1_{g} = (1+{\beta}^2)*{\frac{({precision_g}*{recall_g})}{{\beta}^2*{precision_g}+{recall_g}}},
\end{equation}
{where $precision_g$ and $recall_g$ are precision and recall for group $g$, respectively.} We set {$\beta$} to 1 to balance the recall and precision values.
To get a scalar-valued metric, we take the mean of EF1s of all groups as:
\begin{equation}
\label{eq:metric}
    mEF1 = \frac{\sum_{g=1}^G EF1_{g}}{G}.
\end{equation}

\subsection{{Comparison to Day-time Scene Parsing Methods}}\label{subsec:comp_pri_methd}
\textbf{Compared Methods.} We compare our model with several state-of-the-art scene parsing methods, including SegNet \cite{Badrinarayanan2017SegNet}, FCN-8s \cite{Long2015Fully}, PSPNet \cite{zhao2017pspnet}, BiSeNet \cite{yu2018bisenet}, PSANet \cite{zhao2018psanet}, ESPNet \cite{Mehta2018ESPNet}, DFN
\cite{Yu2018Learning}, CCNet \cite{huang2018ccnet}, \revisionxin{VPLR \cite{Zhu_2019_CVPR} and HANet \cite{choi2020cars}}. We use their released codes and train {them} using the hyper-parameters reported in their papers. Since all of these methods are developed for day-time domain, one naive solution for NTSP is to apply image enhancement to the input night-time images followed by an existing day-time method for scene parsing. Hence, we also evaluate this pre-processing approach using {one of the latest} enhancement {methods} DRHT~\cite{Yang2018Image} in our experiment.

\renewcommand{\arraystretch}{1.5}
\begin{table*}
\centering
\caption{Comparison of our model with state-of-the-art methods on exposure-aware F1 score (EF1), which quantifies the segmentation performance under different exposure levels. From left to right, exposure degree increases from under-exposure to over-exposure. $EF1_g$ denotes different exposure levels (or bins). All models are trained and tested on NightCity. The best results are highlighted in \textbf{bold}.}
\resizebox{\linewidth}{!}{
\begin{tabular}{l|c|c|c|c|c|c|c|c|c|c}
\hline
&\multicolumn{10}{c}{{$EF1_g \uparrow$}}
\\		 \cline{2-11}
Methods &{\rotatebox{0}{[0, 0.1)}}&{{[0.1,0.2)}}&{[0.2, 0.3)} &{[0.3, 0.4)} &{[0.4, 0.5)} &{[0.5, 0.6)} &{[0.6, 0.7)} & {[0.7, 0.8)} &{[0.8, 0.9)} & {[0.9, 1]}  \\
\hline
SegNet \cite{Badrinarayanan2017SegNet}&0.6335 &0.6281&0.6945    &0.7171&0.7132&0.6877 &0.6625&0.6566& 0.6507& 0.6665 \\
FCN-8s \cite{Long2015Fully}&0.7717 & 0.7863 &  0.8069&  0.8189   & 0.8209   & 0.8154 &   0.8027 & 0.8016&0.7952  &  0.8085 \\
PSPNet \cite{zhao2017pspnet}&0.8447&0.8634&0.8739&0.8750&0.8754&0.8739&0.8649& 0.8624&0.8583&0.8882 \\
BiSeNet \cite{yu2018bisenet}&0.8609 &0.8674 &0.8735 &0.8735 &0.8740 &0.8736 &0.8658 &0.8637 &0.8571 &0.9073\\
PSANet \cite{zhao2018psanet}&0.7067 &0.7087 & 0.7252& 0.7228    &0.7128&0.6964  & 0.6778 & 0.6744& 0.6647&0.7252 \\
ESPNet \cite{Mehta2018ESPNet}&0.7860  &  0.8085    &0.8325   & 0.8374   & 0.8399  &  0.8352  &  0.8257    &0.8245 &   0.8178  &  0.8249\\
DFN \cite{Yu2018Learning}& 0.8665 &0.8697& 0.8773&0.8763&0.8762& 0.8762&0.8684    &0.8671& 0.8623&0.9119 \\
CCNet \cite{huang2018ccnet}& 0.8681 &0.8623& 0.8745&0.8737&0.8778& 0.8765&0.8625    &0.8678& 0.8595&0.9109 \\
\revisionxin{VPLR} \cite{Zhu_2019_CVPR}   &0.8612  & 0.8655 & 0.8727 &0.8725 &0.8765 &0.8711 &0.8609 &0.8613 &0.8572 &0.9021  \\
\revisionxin{HANet} \cite{choi2020cars} &0.8656 & 0.8645& 0.8788 &0.8747 &0.8812 &0.8751 &0.8703 &0.8638&0.8632&0.9112  \\
\hline
Ours&\textbf{0.8690}  &  \textbf{0.8751} &   \textbf{0.8819}   & \textbf{0.8819}  &  \textbf{0.8828}  &  \textbf{0.8814}  &  \textbf{0.8739}   & \textbf{0.8738}   & \textbf{0.8689}& \textbf{0.9155}\\
\hline
\end{tabular}}
\vspace{0.05in}
\label{table:sam_ana}
\end{table*}

\begin{figure}[b]
\centering
\includegraphics[width=\linewidth]{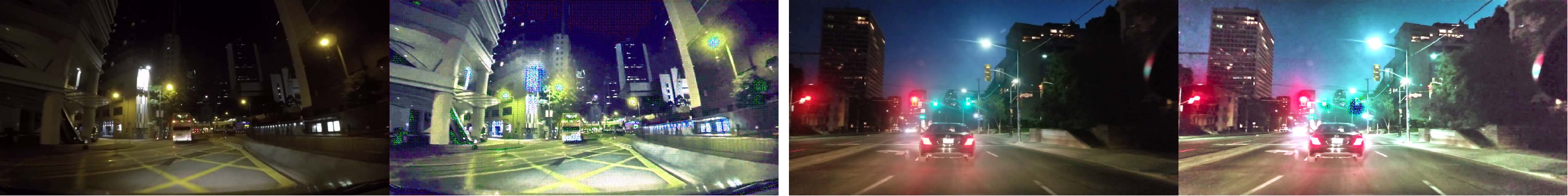}
\footnotesize{\textcolor{white}{}\hspace{3mm} Night \hspace{11mm} DRHT \hspace{10mm} Night \hspace{10mm} DRHT}
\vspace{1mm}
\caption{Examples of night images and the corresponding enhanced images using DRHT~\cite{Yang2018Image}.}
\vspace{-1mm}
\label{fig:enhan_image}
\end{figure}

\textbf{Quantitative Results.} Table \ref{table:result_tabel1} reports the experimental results. We have two observations here:
\begin{enumerate}
\item The existing day-time models trained on Cityscapes achieve poor performances (``C'' columns), which are significantly worse than those of our model trained on NightCity, in terms of mIoU and mEF1.  This shows that NTSP is rather challenging and the state-of-the-art scene parsing methods are not able to handle this problem well.
\item Adding image enhancement as a pre-process does not give obvious performance gains (``C (IE)'' columns).
To review the problems, we use DRHT~\cite{Yang2018Image} to enhance two example night images, as shown in Figure \ref{fig:enhan_image}. We can see that while the under-exposed regions are enhanced, the over-exposure regions get worse. The enhanced images also have very different appearances from day-time ones, causing day-time scene parsing methods to fail.
\end{enumerate}
Results of our model, when trained on our real night-time dataset, show superior performance over these existing methods trained on Cityscapes.

\subsection{{Evaluation of the Proposed NightCity Dataset}}\label{sec:beni_night_data}
To investigate if our dataset can help improve the performance of existing methods, we have conducted {three} experiments.
In the first experiment, we train the existing scene parsing models and our model on NightCity, instead of Cityscapes. In the second experiment, we train all models on Cityscapes and NightCity. {In the third experiment, we compare the performances of existing methods trained on the existing night-time dataset BDD100K-night~\cite{BDD100K} and those trained on our NightCity.}

\begin{figure*}[t]
\begin{minipage}{0.1\linewidth}
\vspace{1mm}\textcolor{white}{0}\\
\vspace{12mm}\small{Input}\\
\vspace{11mm}\small{GT}\\
\vspace{11mm}\small{FCN-8s}\\
\vspace{11mm}\small{PSPNet}\\
\vspace{11mm}\small{BiSeNet}\\
\vspace{11mm}\small{ESPNet}\\
\vspace{11mm}\small{DFN}\\
\vspace{13mm}\small{CCNet}\\
\vspace{13mm}\small{VPLR}\\
\vspace{13mm}\small{HANet}\\
\vspace{1mm}\small{Ours}\\
\end{minipage}
\begin{minipage}{0.9\linewidth}
\includegraphics[width=1.0\linewidth]{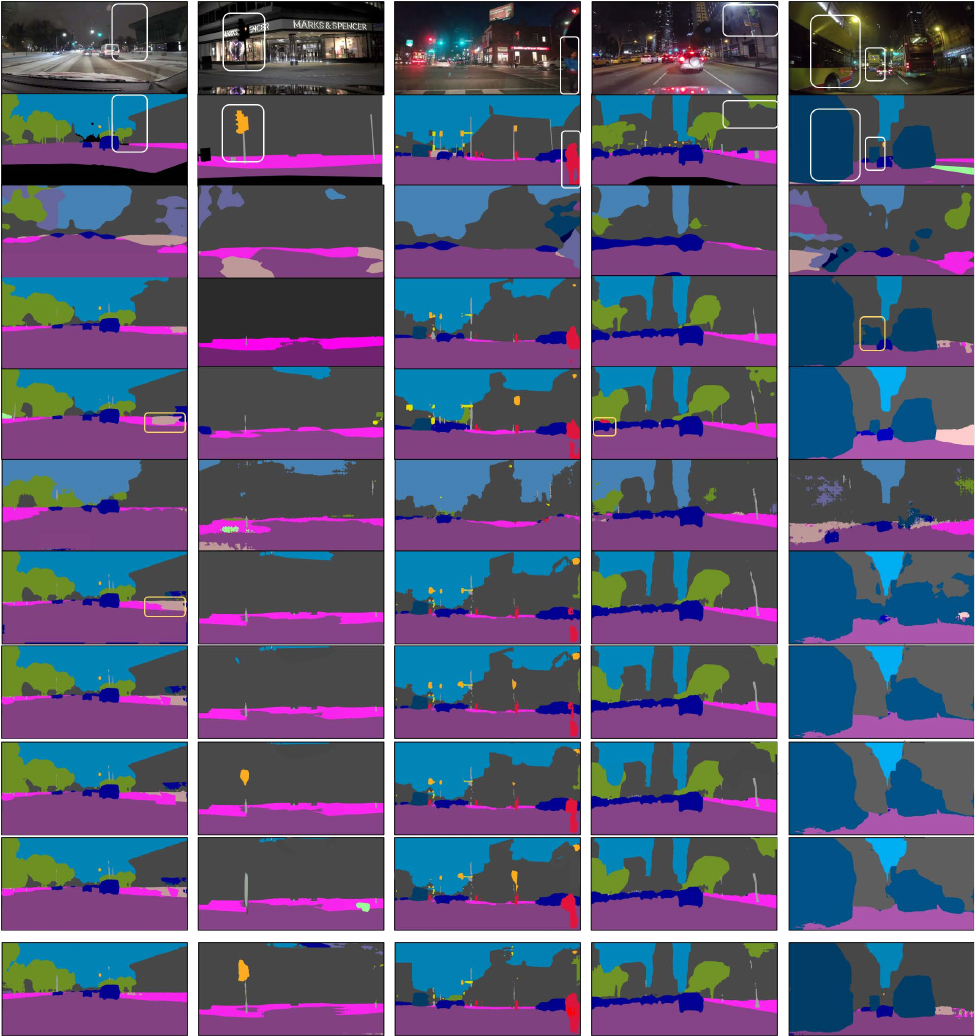}
\end{minipage}
\vspace{2mm}
\caption{Visual comparison of our results with those of the state-of-the-art methods. Our advantages are highlighted by white boxes. A few drawbacks of the other methods are marked by yellow boxes. All the methods are trained on NightCity.}
\label{fig:visual}
\end{figure*}

\textbf{Quantitative Results.} {We report the results of the first and second experiments on the NightCity test set in Table \ref{table:result_tabel1}}.
We have the following observations:
\begin{enumerate}
\item Cityscapes (``C'' columns) vs. NightCity (``N'' columns). We can see that after training on our NightCity, the performances of all existing methods have improved significantly, on both mIoU and mEF1.
This suggests that our real night-time dataset is important to boosting the NTSP performance.

\item 
Although all models have significant performance gains after training on NightCity, our proposed model outperforms all existing methods. This demonstrates that our model, while simple, is very effective in handling night-time images, as compared with other methods. We also note that our model obtains the best mEF1 performance. To reveal how well our model performs under different exposure conditions, we also report the $EF1_g$ values for the 10 exposure groups (i.e., $g=\{1, \dots, 10\}$) in Table \ref{table:sam_ana}. We can see that our model outperforms the other models in all exposure groups, including [0, 0.2] (near under-exposure) and [0.8, 1] (near over-exposure).

\item NightCity (``N'' columns) vs. Cityscapes + NightCity (``C + N'' columns).
We can see that after training on both datasets, some of the existing models have small performance gain, while others have small performance reduction. We believe that the performance gain of some methods may be due to the added context and object information from Cityscapes. On the other hand, the added day-time information may confuse the other methods, causing the performance reduction.
\end{enumerate}

{For the third experiment, we report the results in Table~\ref{table:nightdataset}. As there are currently no large-scale datasets with fine annotations for night-time scene parsing, we verify the effectiveness of the proposed dataset by comparing it with a subset of the BDD100K dataset (with annotations) \cite{BDD100K} (denoted as BDD100K-night here), which contains 314 night-time images for training and 31 for test. We train three methods (i.e., PSPNet \cite{zhao2017pspnet}, CCNet \cite{huang2018ccnet}, and HANet \cite{choi2020cars}) on the BDD100K-night training set and our training set, and test them on the BDD100K-night test set.
The results in Table~\ref{table:nightdataset} show that our dataset can help improve the performances significantly, which demonstrates the effectiveness of our dataset.}

\textbf{Qualitative Results.} Figure \ref{fig:visual} qualitatively compares the results of our model with those of the best-performing six models (according to Table \ref{table:result_tabel1}) on some of the images in NightCity. These images have different degrees of under-/over-exposures, which render them difficult to recognize and segment. However, our model is able to handle them favorably. Particularly,
in the first column, our model can produce more accurate and sharper boundaries on the building segmentation. {In addition, it} also gives more clear and complete sidewalks.
In the second column, the traffic light with its pole is ignored by other models, but successfully recognized by ours.
In the third column, our model gives more intact and clear shape of the pedestrian at the right end.
In the fourth column, our model can detect a very small tree in an under-exposed region. Although BiseNet can also detect the tree, {it generates a false positive segmentation of a person on the left side (the red segment marked by the yellow box).}
In the last column, while both our model and PSPNet are able to segment the buses near the camera well, PSPNet fails to give a correct segmentation of the distant bus (marked by the yellow box).
These results once again {demonstrate} the superior performance of the proposed model on NTSP.

\renewcommand{\arraystretch}{1.2}
\begin{table}[h]
\begin{center}
\caption{Comparison of the existing methods trained on the BDD100K-night dataset and on our NightCity dataset. Methods are evaluated on the BDD100K-night test set.}\label{table:nightdataset}
\begin{small}
{
\begin{tabular}{l| c c| c c}
\hline
\multirow{2}*{Methods} &\multicolumn{2}{c}{BDD100K-night [38]} &\multicolumn{2}{|c}{NightCity-train}\\
\cline{2-5}&mIoU (\%) $\uparrow$ &mEF1  $\uparrow$ &mIoU (\%) $\uparrow$ &mEF1  $\uparrow$ \\

\hline
PSPNet [6]  &  36.8&  0.69 & 51.2&0.76      \\
\hline
CCNet [20]  &  41.4 &0.72 &54.6&0.81 \\
\hline
HANet [21]  & 45.2  &0.74&59.9&0.88\\

\hline
\end{tabular}}
\end{small}
\end{center}
\vspace{-2.5mm}
\end{table}

\begin{figure*}[t]
\centering
\includegraphics[width=\linewidth]{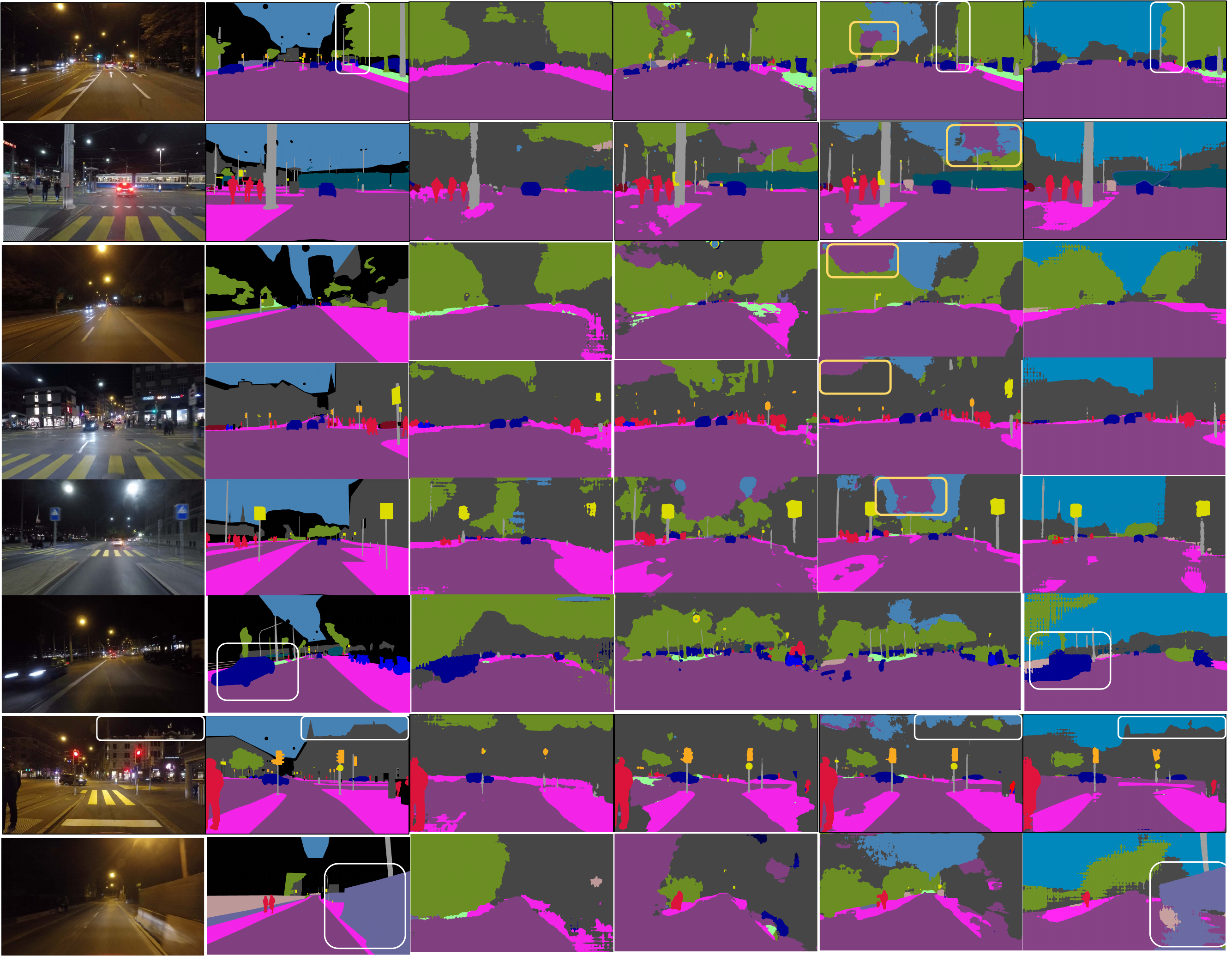}
\footnotesize{Input\hspace{25mm}GT\hspace{17mm}AdaptSegNet\hspace{19mm}DMAda\hspace{19mm}GCMA\hspace{19mm}Ours}
\vspace{1mm}
\caption{Visual comparison of our results with those from AdaptSegNet \cite{tsai2018learning}, DMAda \cite{dai2018dark} and GCMA \cite{GCMA_UAE}. Our advantages are highlighted with white boxes and the limitations of existing methods are highlighted with yellow boxes. Our model can produce more {accurate and robust} segmentation.}
\vspace{-3mm}
\label{fig:other_night}
\end{figure*}

\subsection{{Comparison to Night-time Scene Parsing Methods}}

{{\bf Quantitative Results.} We compare our method to some ad hoc night-time scene parsing methods on one open challenge, i.e., the {\bf Dark Zurich dataset} ~\cite{GCMA_UAE}. This dataset has 151 night-time images for evaluation. Since their labels are only available in their challenge, we submit our results to their challenge to obtain the results.
Methods used for comparison include: GCMA~\cite{GCMA_UAE}, AdaptSegNet~\cite{tsai2018learning}, ADVENT~\cite{vu2019advent}, BDL~\cite{li2019bidirectional}, DMAda~\cite{dai2018dark}, MGCDA~\cite{sakaridis2020map}, DANNet~\cite{wu2021dannet}.
All these methods are domain-adaption-based methods, as there were no large-scale night-time scene parsing datasets with fine annotations for fully-supervised learning.
The results are shown in Table~\ref{table:other_night}, which verifies the superiority of our method.}

{\bf Qualitative Results.} We qualitatively compare our results with the results shown in GCMA \cite{GCMA_UAE}. As they only show results of eight images in their paper (including those in the supplemental), we therefore visually compare our results with their results of the eight images. Their results were from {three NTSP methods, AdaptSegNet \cite{tsai2018learning}, DMAda \cite{dai2018dark}, and GCMA \cite{GCMA_UAE}.}
{These three compared methods are the same as those given in GCMA \cite{GCMA_UAE}. As shown in the top four rows of Figure \ref{fig:other_night}, other methods mistakenly recognize a part of the sky as road (yellow boxes), while our model}
can produce the very clear boundary of the tree (white boxes).
{In the fifth row, our model can detect a clear shape of the car.}
In the {seventh} row, our model produces a much cleaner building (white box), compared with another three methods. {In the last row, only our model can identify the wall.}
These visual results show that our model can produce higher-quality segmentation maps under extreme lighting conditions.

\renewcommand{\arraystretch}{1.2}
\begin{table}[tb]
\begin{center}
\caption{{Comparison  of  our  model  with  state-of-the-art  methods  on Dark Zurich-test \cite{GCMA_UAE}. The best result is highlighted in {\bf BOLD}.}}\label{table:other_night}
\begin{small}
{
\begin{tabular}{c c c c }
\hline 		
Methods  &Venue\&Year & mIoU (\%) $\uparrow$   \\
\hline
AdaptSegNet \cite{tsai2018learning} &CVPR 2018&  30.4   \\
\hline
ADVENT \cite{vu2019advent} & CVPR 2019   &29.7\\
\hline
BDL \cite{li2019bidirectional}  & CVPR 2019      &30.8\\
\hline
DMAda \cite{dai2018dark} &ITSC 2018 &  32.1   \\
\hline
GCMA \cite{GCMA_UAE} &ICCV 2019  & 42.0  \\
\hline
MGCDA  \cite{sakaridis2020map} & TPAMI 2021           & 42.5\\
\hline
DANNet (DeepLab-v2) \cite{wu2021dannet} & CVPR 2021 &42.5\\
\hline
DANNet (RefineNet) \cite{wu2021dannet}  & CVPR 2021  &44.3\\
\hline
DANNet (PSPNet)  \cite{wu2021dannet}  & CVPR 2021   &45.2\\
\hline
Ours & - &\textbf{45.4}    \\
\hline
\end{tabular}}
\end{small}
\end{center}
\vspace{-2.5mm}
\end{table}

\renewcommand{\arraystretch}{1.2}
\begin{table}[h]
\begin{center}
\caption{Comparison  of  our  model  with  state-of-the-art  methods  on  day-time images (the validation set of Cityscapes) using mIoU.  The best and second best results are marked in {\bf BOLD} and {\color{blue}{blue}}, respectively.
}\label{table:citysc}
\begin{small}
{
\begin{tabular}{c  c  c c }
\hline 		
 Methods & Venue\&Year & mIoU (\%) $\uparrow$   \\
\hline
FCN-8s \cite{Long2015Fully} &CVPR 2015 &62.2    \\
\hline
SegNet \cite{Badrinarayanan2017SegNet} &TPAMI 2017  & 51.5    \\
\hline
PSPNet \cite{zhao2017pspnet} &CVPR 2017 &75.3    \\
\hline
BiSeNet \cite{yu2018bisenet}&ECCV 2018 & 39.6   \\
\hline
PSANet \cite{zhao2018psanet}&ECCV 2018 &41.2   \\
\hline
ESPNet \cite{Mehta2018ESPNet}&ECCV 2018 & 40.6   \\
\hline
DFN \cite{Yu2018Learning}&CVPR 2018 &75.8    \\
\hline
CCNet \cite{huang2018ccnet}&ICCV 2019 &\textbf{77.1}   \\
\hline
 CGNet \cite{Wu2021seg}& TIP 2021 &64.8 \\
\hline
DSD \cite{9444191}& TIP 2021  &72.3\\
\hline
MagNet \cite{huynh2021progressive}& CVPR 2021  &65.6\\
\hline
Ours & - &\textcolor{blue}{76.9}    \\
\hline
\end{tabular}}
\end{small}
\end{center}
\vspace{-2.5mm}
\end{table}

\subsection{Model Analysis}

{\bf Model Robustness on Day-time Dataset.} {We evaluate our model on the existing day-time dataset Cityscapes to show its robustness. We provide the mIoU results of different methods trained on both Cityscapes and NightCity, and evaluated on the validation set of Cityscapes. The results are reported in Table \ref{table:citysc}.}
{We can see that although our method is not specifically designed for day-time scene parsing, it can still achieve the second-best performance compared with the state-of-the-art methods on day-time images. }

{\bf Ablation Study.} To investigate the necessity of using the learned exposure features to guide the segmentation, we run an ablation study to compare our model against its four ablated alternatives: (1) remove the exposure guidance layer from our model (``$w/o EGL$''); (2) remove the entire exposure stream from our model (``$w/o ES$''); (3) replace the attention operation in Eq.~\ref{eq:attention} with concatenation (Concat); and (4) with summation (Sum).
We include alternatives (3) and (4) as they are straightforward ways of combining multiple features.
Table \ref{table:exposure_stream} shows the results. We can see that when the exposure guidance layers are excluded, the performance of our model drops significantly, which confirms the importance of our exposure guidance. {If we remove the exposure stream (``$w/o~ES$''), the performance drops further, as there are no mechanisms for handling the over- and under-exposed regions. This suggests that learning to explicitly predict exposure information can help learn useful features for NTSP,} Finally, we can see that using concatenation and sum operations for fusing the exposure information can produce slightly better results.
However, they still perform much worse than ours, which demonstrates the effectiveness of our attention operations in EGLs for exposure guidance, {as the attention mechanism allows our method to aggregate different contextual information for segmentation according to the exposure-based attention map}.

\renewcommand{\arraystretch}{1.2}
\begin{table}[h]
\begin{center}
\caption{
Ablation study. We compare our model (Ours) with its ablated versions: without exposure guidance layers (w/o EGL), without the exposure stream (w/o ES), and replacing the attention operations with concatenation (Concat) and sum (Sum).}\label{table:exposure_stream}
\begin{small}
{
\begin{tabular}{c c c c c c}
\hline 		
  & w/o EGL & w/o ES & Concat & Sum &Ours\\
\hline
mIoU (\%) $\uparrow$& 41.6 & 40.1  &42.5&43.1    &51.8  \\
\hline
mEF1 $\uparrow$&0.83 &0.82 &0.84 & 0.82  &0.88  \\
\hline
\end{tabular}}
\end{small}
\end{center}
\end{table}

{\bf Strategies of Using the Exposure Map.} To incorporate exposure information into a scene parsing model, one simple strategy is to directly use an exposure map as an additional input to the model or as an attention map to fuse the intermediate features. To justify the advantage of our network design over these straightforward solutions, we compare our model with two baselines: (1) we use the exposure map as an additional input to our segmentation stream, by concatenating the exposure map with the RGB image, and remove the exposure stream (denoted as ``as Extra Input'');
(2) we replace the attention map $W_r$ in Eq.~\ref{eq:attention} with the exposure map and remove the exposure stream (denoted as ``as Attention Map'').

\begin{figure}[!htbp]
\vspace{-2mm}
\centering
\includegraphics[width = \linewidth]{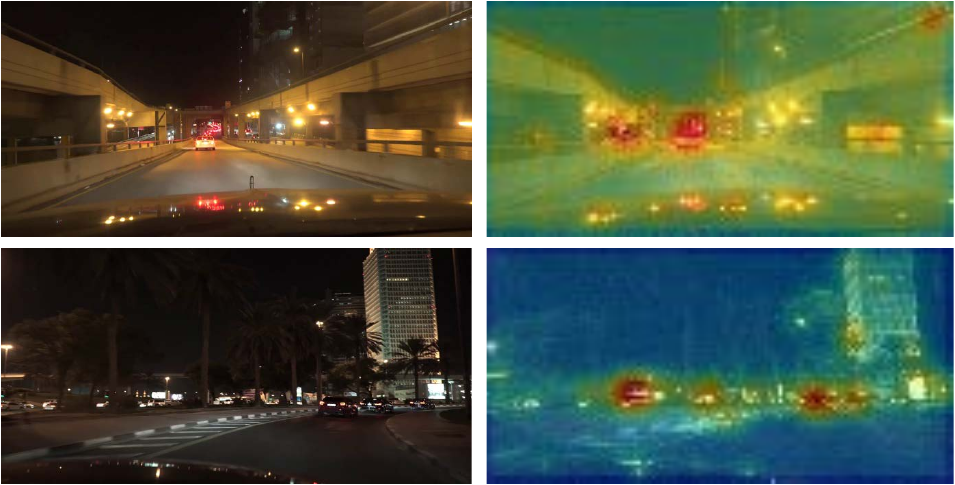}
\caption{Visualization of the attention maps learned by our guidance of exposure layer.}
\label{fig:exposure_atten}
\vspace{-4mm}
\end{figure}

As shown in Table \ref{table:exposure_map}, our model outperforms the two \revisionxin{ablated models} by a large margin.
{If we feed the exposure map as input to the network, the network can only be supervised by the semantic labels.
Since the exposure map does not contain much semantic information, its contribution to semantic segmentation prediction would not be modeled by the network.
Table~\ref{table:exposure_map} shows that using the exposure map as input, instead of as the supervision signal, causes a significant performance drop from 51.8 to 35.6 (in terms of mIoU).
In addition, by comparing Table \ref{table:exposure_map} to Table \ref{table:exposure_stream}, we can see that using the exposure map as input produces a worse performance (35.6 mIoU in Table~\ref{table:exposure_map}) than the ablated model of removing the exposure stream ``$w/o~ES$'' (40.1 mIoU in Table \ref{table:exposure_stream}).
This suggests that the network may mistakenly use the exposure map as a weight map, which pays more attention to the over-exposed regions and less attention to the under-exposed regions.
Table~\ref{table:exposure_map} further shows that directly using the exposure map as an attention map produces a similar result (40.2 mIoU in Table \ref{table:exposure_map}) as the ablated model ``$w/o~ES$'' (40.1 mIoU in Table \ref{table:exposure_stream}).
This demonstrates that exposure information may not help in such an attentive manner.}

{Figure \ref{fig:exposure_atten} shows examples of visualization of the attention maps learned by the exposure guidance layer. The first row shows that our model gives more attention to distant cars with lots of headlights (over-exposure) and the second row shows that our model pays more attention to the under-exposed regions.}
In contrast to the direct usage of exposure map as attention map that network would pay more attention to over-exposed regions,
our model can learn to adaptively attend to both over- and under-exposed regions as all of them are crucial to segmentation performance of night-time images.

{These experiments verify the effectiveness of our method of exploiting the exposure information, by first learning the exposure representation via the image-to-exposure translation step and then learning the fusion of semantic and exposure information via the exposure guidance layer.}

\renewcommand{\arraystretch}{1.2}
\begin{table}[h]
\begin{center}
\caption{
Comparison with two {ablated models} of using the exposure map: using it as an additional input (as Extra Input) and as an attention map (as Attention Map).}\label{table:exposure_map}
\begin{small}
{
\begin{tabular}{c c c c}
\hline 		
  & {as Extra Input} & {as Attention Map}  & Ours\\
\hline
mIoU (\%) $\uparrow$&35.6  & 40.2 &51.8\\
\hline
mEF1 $\uparrow$& 0.73&0.76   &0.88  \\
\hline
\end{tabular}}
\end{small}
\end{center}
\vspace{-2.5mm}
\end{table}

\begin{figure}[h]
\centering
\includegraphics[width=\linewidth]{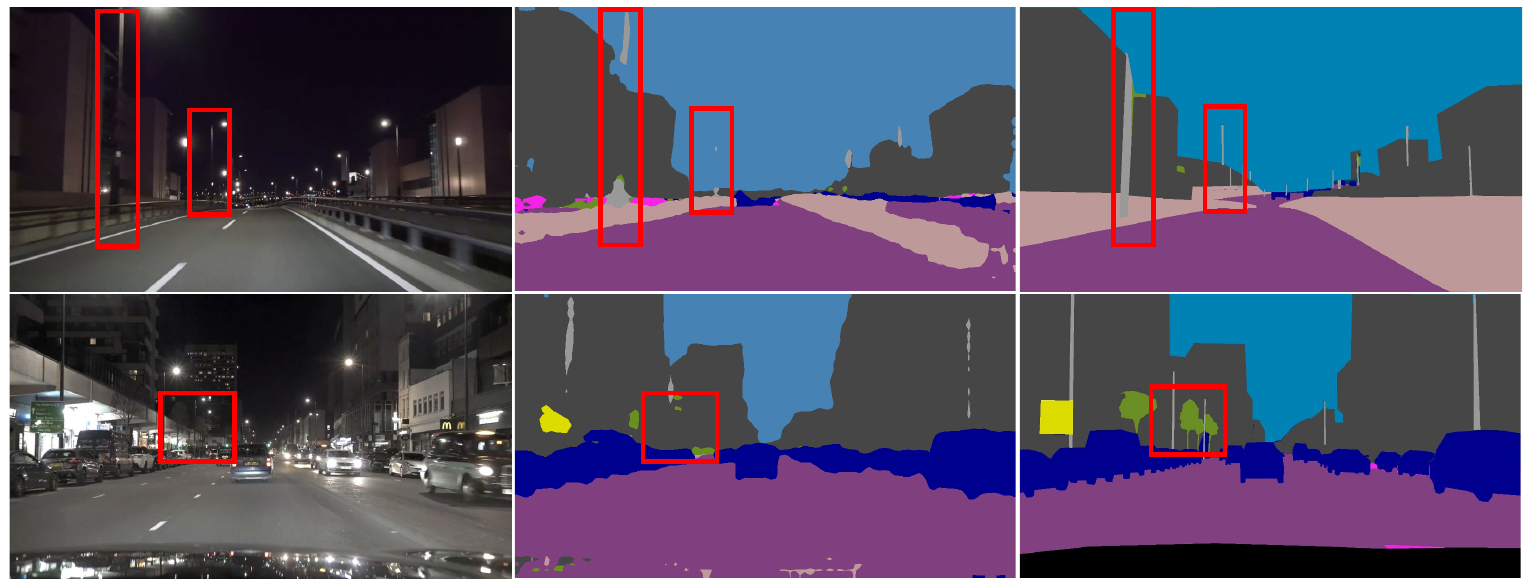}
\footnotesize{Input \hspace{20mm} Ours \hspace{20mm} GT}
\vspace{1mm}
\caption{Failure cases. Our model may fail to detect objects that appear in large under-exposed regions, e.g., thin poles (top row) and trees (bottom row).}
\vspace{-0.5cm}
\label{fig:fail_case}
\end{figure}

\section{Conclusion}
In this paper, we have addressed the night-time scene parsing (NTSP) problem. To this end, we have proposed a large dataset of real night-time images with fine semantic annotations for training and benchmarking.
We have also proposed a two-stream framework especially designed to address the NTSP problem, which explicitly learns exposure features to augment the scene parsing process.
Our results show that the proposed dataset can benefit existing scene parsing methods when applied to night-time scenes. We have also demonstrated that our proposed model trained on our dataset outperforms all existing methods, yielding state-of-the-art performance.

Although we have demonstrated the effectiveness of our model on night-time scenes, our model may fail in some extremely challenging situations, e.g., if an under-exposure region is large. Figure \ref{fig:fail_case} shows two failure examples of our model, in which it fails to detect the thin poles (top row) and the trees (bottom row). All these objects are located in large under-exposed regions and are visually difficult to identify even for human.
As a future work, we would like to consider using the raw data from the camera to address this problem.

\section*{Acknowledgments}
We thank the reviewers for insightful comments and constructive suggestions. This work is partially supported by the National Key Research and Development Program of China {\small (No. 2019YFC1521104)}, National Natural Science Foundation of China {\small (No. 61972157)}, Zhejiang Lab {\small (No. 2020NB0AB01)}, Shanghai Municipal Science and Technology Major Project {\small (2021SHZDZX0102)}, Shanghai Science and Technology Commission  {\small (21511101200)}, a GRF from the Hong Kong Research Grants Council {\small (RGC Ref: 11205620)}, and a SRG from City University of Hong Kong {\small (Ref: 7005674)}. Xin Tan is supported by the Postgraduate Studentship (Mainland Schemes) from City University of Hong Kong.



\ifCLASSOPTIONcaptionsoff
  \newpage
\fi

\bibliographystyle{IEEEtran}
\bibliography{egbib}




\end{document}